\documentclass{article}

\PassOptionsToPackage{numbers, compress}{natbib}
\usepackage{cite}
\usepackage[pagebackref]{hyperref}

\usepackage[preprint]{neurips_2024}

\usepackage[utf8]{inputenc}
\usepackage[T1]{fontenc}
\usepackage{hyperref}
\usepackage{url}
\usepackage{booktabs}
\usepackage{amsfonts}
\usepackage{nicefrac}
\usepackage{microtype}
\usepackage{xcolor}

\usepackage{csquotes}
\usepackage{CJKutf8}
\usepackage{graphicx}
\usepackage[most]{tcolorbox}
\usepackage{cleveref}
\usepackage{multirow}

\title{Formal Mathematical Reasoning: \\A New Frontier in AI}

\author{%
    Kaiyu Yang$^{1}$, Gabriel Poesia$^{2}$, Jingxuan He$^{3}$, \\
    \textbf{Wenda Li$^{4}$, Kristin Lauter$^{1}$, Swarat Chaudhuri$^{5}$, Dawn Song$^{3}$}\\
    $^1$Meta FAIR, $^2$Stanford University, $^3$UC Berkeley, $^4$University of Edinburgh, $^5$UT Austin
}

\newcommand{\smallsec}[1]{\vspace{0mm}\paragraph{#1.}}

\begin{document}

\maketitle

\begin{abstract}
AI for Mathematics (AI4Math) is not only intriguing intellectually but also crucial for AI-driven discovery in science, engineering, and beyond. Extensive efforts on AI4Math have mirrored techniques in NLP, in particular, training large language models on carefully curated math datasets in text form. As a complementary yet less explored avenue, \emph{formal mathematical reasoning} is grounded in formal systems such as proof assistants, which can verify the correctness of reasoning and provide automatic feedback. In this position paper, we advocate for formal mathematical reasoning and argue that it is indispensable for advancing AI4Math to the next level. In recent years, we have seen steady progress in using AI to perform formal reasoning, including core tasks such as theorem proving and autoformalization, as well as emerging applications such as verifiable generation of code and hardware designs. However, significant challenges remain to be solved for AI to truly master mathematics and achieve broader impact. We summarize existing progress, discuss open challenges, and envision critical milestones to measure future success. At this inflection point for formal mathematical reasoning, we call on the research community to come together to drive transformative advancements in this field.
\end{abstract}

\section{Introduction}

Since the early days of AI, researchers have dreamed of building AI systems that can automate mathematical reasoning. The first AI program in history was Newell and Simon's Logic Theorist~\cite{newell1956logic}, a theorem proving system that could prove 38 theorems in \emph{Principia Mathematica}~\cite{whitehead1927principia}. In the decades since then, the center of AI has shifted from symbolic methods to machine learning, and a new field of statistical AI for mathematics (AI4Math) has emerged. One appeal of the field is that mathematical problems are a proxy for a broad array of reasoning and planning tasks. Another attraction is that math plays a foundational role in quantitative disciplines, so AI4Math has the potential to revolutionize AI for science, engineering, and beyond.
For these reasons, designers of large language models (LLMs)~\cite{openai2023gpt4,touvron2023llama} have frequently highlighted LLMs' success in math problems, and there have also been efforts to build AI systems that outperform humans at math competitions~\cite{alphageometry,aimo2024,alphaproof}.

Given the importance of AI4Math, substantial research has been dedicated to developing math LLMs, using techniques borrowed from natural language processing (NLP). A common approach is to continue pretraining LLMs on math data, such as arXiv papers and web pages from MathOverflow, and then finetune the model on curated datasets of math problems with detailed, step-by-step solutions. We call this the ``informal'' approach to distinguish it from the formal approach that will be introduced later (Sec.~\ref{sec:background}). Just like LLMs in general, math LLMs have a simple recipe, but the secret sauce is often data curation~\cite{yu2024metamath,yue2024mammoth,azerbayevllemma,shao2024deepseekmath}. Carefully curated training data plus inference-time techniques, including chain-of-thought prompting~\cite{wei2022chain}, self-consistency~\cite{wang2023self}, and tool use~\cite{gou2024tora}, have led to remarkable success on widely used benchmarks such as GSM8K~\cite{cobbe2021training} and MATH~\cite{hendrycks2021measuring}, as well as in the AIMO Progress Prize~\cite{aimo2024}. However, at the time of writing, the success of the informal approach has been mostly limited to high school math not exceeding the AIME level.\footnote{Some of the most difficult problems in MATH and the AIMO Progress Prize are at the level of AIME (American Invitational Mathematics Examination). OpenAI o1~\cite{o1} was also evaluated on AIME problems.} This raises a key question: \emph{How far can we go by scaling up the informal approach? Will it enable math LLMs to solve more challenging competition problems (e.g., IMO, International Mathematical Olympiad) or even problems in mathematical research?}

Moving from high school to more advanced mathematics, the informal approach faces challenges that are hard to resolve by merely scaling up the training. First, training math LLMs requires high-quality data, which is scarce in advanced mathematics. For novel research math problems, it is infeasible to find solutions to similar problems on the Internet or manually annotate the data on a large scale. Without scaling up the data, we cannot fully benefit from the scaling laws for LLMs~\cite{hoffmann2022training,zhang2024scaling}. Second, solutions to many advanced problems are not numbers that can be evaluated by comparing them with the ground truth. Instead, they carry out a chain of intricate reasoning steps, e.g., a proof. LLMs are notorious for hallucinating seemingly valid reasoning steps, making it challenging to evaluate the correctness of model output or collect useful feedback for learning. These challenges are difficult to address by scaling up the informal approach during training. \emph{If training-time scaling is not enough, what else do we need?} One emerging direction, exemplified by OpenAI o1~\cite{o1}, is to scale up the informal approach during inference, potentially combining search with neural verifiers to mitigate hallucinated reasoning~\cite{cobbe2021training}. While this approach has gained traction, its effectiveness on advanced mathematical problems is an open question. In this position paper, we focus on a complementary approach that is less explored: \textbf{formal mathematical reasoning.}

We consider formal mathematical reasoning broadly as \emph{mathematical reasoning grounded in formal systems}, including but not limited to first/higher-order logic~\cite{nipkow2002isabelle}, dependent type theory~\cite{barras1997coq}, and computer programs annotated with formal specifications~\cite{leino2010dafny}. Such formal systems provide environments that can verify the model's reasoning and provide automatic feedback. They stand apart from the ``tools'' used by modern LLMs~\cite{schick2024toolformer} in their ability to model the provable truth or falsity of a broad class of propositions. The feedback provided by such systems can mitigate data scarcity; also, such systems enable rigorous test-time checks that resist hallucination. In contrast, \emph{informal mathematics} refers to math text commonly found in textbooks, research papers, and online math forums. Informal math interleaves natural language with symbols (e.g., \LaTeX), but these symbols do not have a self-contained formal semantics, instead relying on informal text to convey significant parts of their meaning. 

AlphaProof~\cite{alphaproof} and AlphaGeometry~\cite{alphageometry} are two prominent examples of the success of this idea. Before these systems, there were many failed attempts to use LLMs to solve olympiad-level math problems. The key differentiator in the aforementioned systems is the principled use of symbolic representations and proof-checking frameworks. The symbolic components (Lean~\cite{de2015lean,moura2021lean} for AlphaProof; a domain-specific geometry system for AlphaGeometry) are used to execute a neural network's reasoning steps and generate high-quality synthetic data, leading to unprecedented mathematical reasoning abilities.  

AlphaProof and AlphaGeometry follow in the footsteps of a broader literature on the synergistic use of formal methods and machine learning in mathematical tasks \cite{urban2011malecop,kaliszyk2018reinforcement,gauthier2021tactictoe,irving2016deepmath,holstep,loos2017deep,huang2019gamepad,yang2019learning}. This literature includes research on neural theorem proving, i.e., generating formal proofs given formal theorem statements~\cite{polu2020generative,yang2023leandojo,thakur2024language}, and autoformalization, i.e., automatically translating informal mathematics into formal mathematics~\cite{wu2022autoformalization}. The advent of LLMs has significantly accelerated research in this area. For example, autoformalization was long hampered by the lack of aligned informal-formal pairs for finetuning. LLMs can mitigate this problem by either synthesizing the data~\cite{jiang2023multilingual} or performing autoformalization without finetuning~\cite{wu2022autoformalization}. As a result, we are starting to realize autoformalization's potential in bootstrapping the capability of neural theorem provers~\cite{xin2024deepseek}. LLMs are also powerful tools for theorem proving; in particular, recent approaches have exploited LLMs to predict proof steps and fix buggy proofs without explicit training on formal proof data~\cite{thakur2024language,first2023baldur}. 

The research infrastructure around LLMs and formal reasoning is rapidly maturing. Lean~\cite{de2015lean,moura2021lean}---a language for writing formal proofs---has gained popularity among mathematicians, leading to formalized research mathematics~\cite{liquid2022} and general-purpose mathematical libraries~\cite{mathlib}.
There are now multiple frameworks~\cite{yang2023leandojo,thakur2024language} that support the interaction between LLMs and Lean. 
These frameworks allow the extraction of training data from human-written formal proofs, as well as theorem proving via interaction with the formal environment. 
Multilingual infrastructures for proof languages like Coq~\cite{barras1997coq} and Isabelle~\cite{nipkow2002isabelle} in addition to Lean are also beginning to be built~\cite{thakur2024language}. Finally, LLMs have been used to assist human mathematicians in writing formal proofs~\cite{song2024towards}, potentially initiating a data flywheel where growing human-written formal math data leads to more capable LLMs, which in turn eases the creation of more data.

The emerging opportunities of AI for formal mathematical reasoning have led to booming research activities. As a recent survey~\cite{li2024dl4tp} shows, the number of publications in this field almost doubled in 2023 and is likely to double again in 2024. By combining autoformlization with reinforcement learning, AlphaProof~\cite{alphaproof} is the first AI to achieve the level of silver medal in IMO. 
Developments in this field also have immediate applications in formal verification~\cite{appel2011verified,klein2009sel4,leroy2016compcert,hawblitzel2014ironclad}, a core computer science problem that has traditionally been among the foremost applications of formal mathematics. While formal verification can lead to software and hardware systems that are exceedingly robust and secure, it has historically been too costly to deploy in all but the most safety-critical applications. 
AI can drastically reduce this cost by substantially automating the formalization and proof effort needed to formally certify complex systems. 
This can lead to a future in which mass-produced software and hardware systems are far more robust than they are today. 

For all these reasons, we believe \textbf{AI-based formal mathematical reasoning has reached an inflection point}, with significant progress to come in the next few years. However, substantial work remains to be done. This position paper maps out open challenges in data and algorithms, as well as potential routes for future progress. It is not meant to be a comprehensive survey but to provide perspectives on where the field may go next and call on the community to unite to accelerate the progress. While we celebrate the promise of formal mathematical reasoning, it should be seen as complementary to the informal approach. For example, future models could combine natural language reasoning with autoformalization to solve informal problems rigorously (Sec.~\ref{subsec:verified-natural-language} and \ref{sec:milestone-natural-language}). 

The remainder of this paper is organized as follows: Sec.~\ref{sec:background} discusses the informal approach in detail and introduces formal mathematical reasoning. Sec.~\ref{sec:progress} reviews recent progress in using AI to reason formally. Sec.~\ref{sec:roadmap} explores open challenges and future directions, and Sec.~\ref{sec:milestones} proposes milestones for measuring AI's capabilities in formal mathematical reasoning.

\section{AI for Mathematics (AI4Math) and the Formal Turn}
\label{sec:background}

Mathematical reasoning is a challenge at the frontier of AI research. In this section, we begin by examining the informal approach to AI4Math and its limitations. Then, we introduce formal mathematical reasoning as a promising path for advancing AI4Math.

\subsection{State-of-the-art Math LLMs and Their Limitations}
\label{sec:mathllm}

\begin{figure}[ht]
  \centering
  \includegraphics[width=1.0\linewidth]{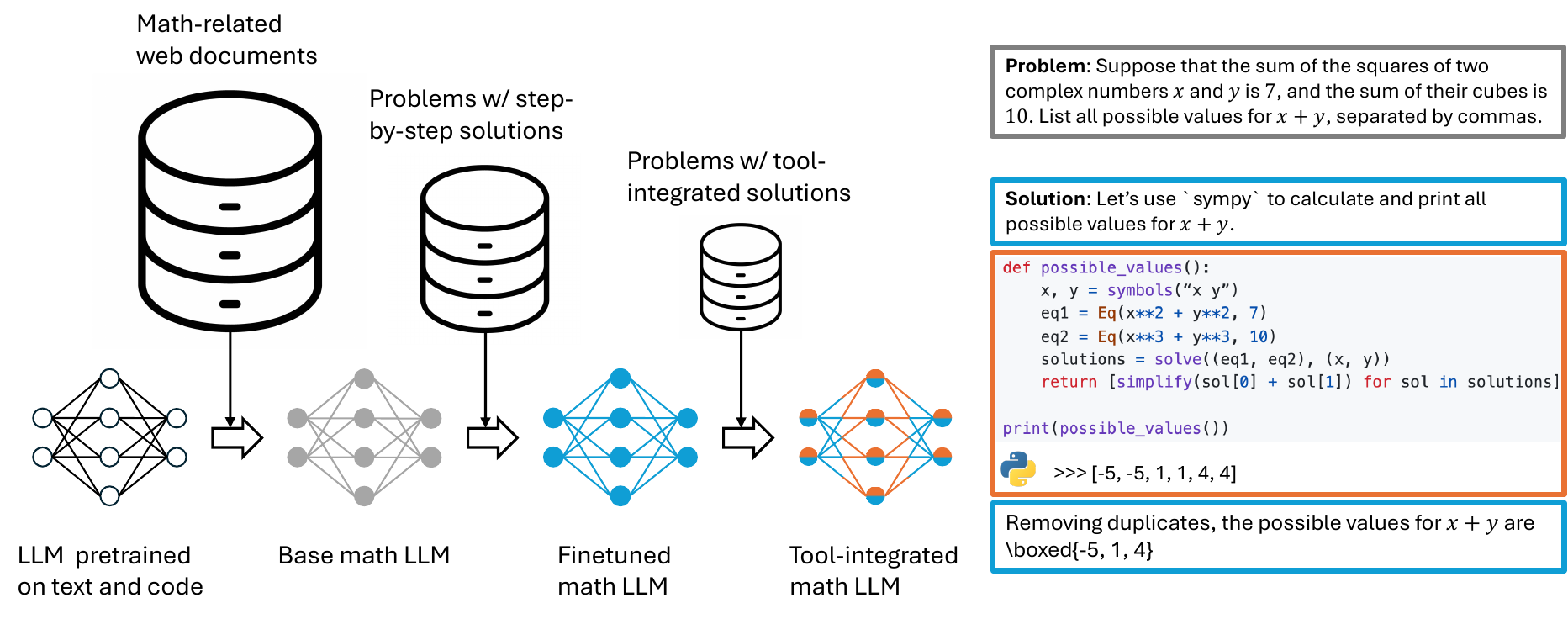}
  \caption{State-of-the-art math LLMs such as NuminaMath~\cite{numina} typically undergo three stages: math pretraining, finetuning on step-by-step solutions, and further finetuning on tool-integrated solutions that interleave natural language reasoning with Python tool invocation.}
  \label{fig:mathLLM}
\end{figure}

\smallsec{A Case Study of NuminaMath}
NuminaMath~\cite{numina} is a math LLM that won the first AIMO Progress Prize in July 2024, successfully solving 29 out of 50 test problems. The test problems were intermediate-level high school math problems newly created and kept private before the evaluation. Therefore, they have very little risk of data contamination compared to public benchmarks such as GSM8K~\cite{cobbe2021training} and MATH~\cite{hendrycks2021measuring}. NuminaMath is an excellent example of state-of-the-art math LLMs, as it encompasses many key ingredients such as math pretraining~\cite{lewkowycz2022solving,ying2024internlm,azerbayevllemma,paster2024openwebmath}, finetuning~\cite{yu2024metamath,yue2024mammoth}, and tool-integrated reasoning~\cite{gou2024tora,yin2024mumath} (Fig.~\ref{fig:mathLLM}). Next, we use NuminaMath as an example to elaborate on each ingredient, highlighting the critical role of \emph{data}.\footnote{OpenAI o1~\cite{o1} demonstrated impressive math capabilities on AIME. It has likely incorporated additional techniques beyond those in Sec.~\ref{sec:mathllm}, but public information on its inner workings remains limited.}

\begin{enumerate}
    \item \emph{Math pretraining} (Fig.~\ref{fig:mathLLM} \emph{Left}): Starting from a generic LLM (or a coding LLM such as Code Llama~\cite{roziere2023code}), one can continue to pretrain the model on a large corpus of math-related documents from the web. The result is referred to as the base math LLM. NuminaMath and other top contestants in the AIMO Progress Prize unanimously adopted DeepSeekMath-Base 7B~\cite{shao2024deepseekmath} as the base math LLM. Critical to DeepSeekMath's success is \emph{data}. To retrieve high-quality math documents from Common Crawl, the authors of DeepSeekMath engineered a data selection pipeline that combined automatic filtering and manual annotation.
    \item \emph{Finetuning on step-by-step solutions} (Fig.~\ref{fig:mathLLM} \emph{Middle}): The base math LLM has been exposed to a vast amount of mathematical content during pretraining, but it is not yet capable of generating well-structured solutions to math problems. To align the model with problem solving, one can finetune it on a carefully curated dataset consisting of math problems with detailed, step-by-step solutions, e.g., in the form of chain-of-thought~\cite{wei2022chain}. This dataset can be constructed by preprocessing and combining heterogeneous sources of problems, e.g., online forums, high school exams, math competitions, problem sets, and manual annotations. The problems and solutions are then augmented and reformatted by LLMs like GPT-4~\cite{openai2023gpt4}. NuminaMath, for example, has constructed a large dataset of 860K problems and solutions covering high school and competition math~\cite{NuminaMath}.
    \item \emph{Tool-integrated reasoning} (Fig.~\ref{fig:mathLLM} \emph{Right}): Finetuned math LLMs have acquired general problem-solving skills, but they may still struggle with precise calculation (e.g., $162 \times 731$) and symbol manipulation (e.g., expanding $(x+2)^{8}$ into powers of $x$). A simple solution is to outsource these operations to external tools such as SymPy~\cite{meurer2017sympy}. NuminaMath performs tool-integrated reasoning that interleaves reasoning in natural language with tool invocation in Python. The key is, again, \emph{data}. The model is finetuned on tool-integrated solutions consisting of natural language combined with tool invocation trajectories. NuminaMath follows the approaches in ToRA~\cite{gou2024tora} and MuMath-Code~\cite{yin2024mumath} to collect this dataset of math problems with tool-integrated solutions.
\end{enumerate}

\smallsec{Data Scarcity}
The NuminaMath team summarized: ``Good data is all you need''~\cite{numina}. Indeed, training data plays a pivotal role throughout all ingredients of the informal approach. As a result, the success of this approach has been limited to domains where abundant high-quality data can be obtained at low costs. For pre-college math, it is relatively easy to collect problems and solutions on the Internet or annotate them manually. However, it is difficult to extend the informal approach to data-scarce domains such as advanced mathematics.

Advanced mathematics forms the foundation of numerous scientific disciplines. For example, climate modeling depends on partial differential equations. To unlock AI's full potential in scientific discovery, it must be able to learn and apply advanced mathematics. Moreover, the long-term goal of developing human-level AI mathematicians requires AI to handle novel aspects of mathematics. Novelty, by definition, implies difficulty in collecting in-distribution training data. Therefore, moving forward, we see data scarcity as a major roadblock to the informal approach to AI4Math.

\smallsec{Lack of Correctness Verifiability}
Besides data scarcity, another challenge is in evaluation, which is essential for AI to make measurable progress. Existing math LLMs are evaluated on benchmarks such as GSM8K and MATH, which consist of math problems whose solution is a single number (or expression). Therefore, evaluation can be done easily, by comparing the model-generated number against the ground truth. While this approach is suitable for pre-college mathematics, it is not directly applicable to more advanced mathematics. Recent work has attempted to adapt this evaluation framework for advanced problems by restricting to problems that have numeric solutions~\cite{glazer2024frontier}. However, this deviates from common practice in mathematical research. After all, almost all AIME problems have numeric solutions, but none of the Millennium Prize Problems do.

Rather than restricting to numeric solutions, advanced mathematics frequently deals with abstract conjectures and proofs. Verifying proofs can be a daunting task, even for experienced mathematicians. This is evident in the lengthy review process for mathematical manuscripts and the controversies surrounding certain proofs, such as the proof of the abc conjecture~\cite{abc}. The situation becomes even more complicated when LLMs are used to generate proofs, as they are known to hallucinate plausibly.

To address the verification difficulty, researchers have explored self-verification or self-correction. These approaches use LLMs to detect and correct reasoning errors in proofs, based on the assumption that verifying proofs may be easier than generating them. While self-verification has been studied extensively, the results have been mixed~\cite{tyen2023llms,miao2024selfcheck,ling2024deductive}. A growing body of work suggests that current LLMs struggle with \emph{intrinsic self-verification}, i.e., verifying their own generations without relying on external feedback~\cite{huang2024large,stechly2024self,gou2024critic,zheng2024processbench,gu2024counterfeit}. Therefore, it is still an open question whether rigorous proof verification can be achieved within the informal approach.

\subsection{AI for Formal Mathematical Reasoning}
\label{sec:formal_mathematical_reasoning}

\smallsec{From Informal to Formal}
Due to the challenges in data and evaluation, it is difficult to directly extend the informal approach to advanced mathematics. \emph{Formal mathematical reasoning} helps address these challenges. In this paper, it refers to mathematical reasoning grounded in formal systems, which have a syntax for well-formed formulas and can perform reasoning by manipulating formulas following a set of well-defined inference rules. Examples of formal systems include axiomatic set theory~\cite{trybulec1989tarski,megill2019metamath}, higher-order logic~\cite{gordon2000lcf,harrison1995metatheory,paulson1988formulation}, and dependent type theory~\cite{martin1984intuitionistic,coquand:inria-00076024,carneiro2024lean4lean}. They are widely used in mathematics and computer programming. In math, they can express axioms, theorems, and proofs. In programming, they are used to specify programs and reason about semantics. The connection between mathematical proofs and computer programs is deepened by theoretical results such as the Curry–Howard correspondence~\cite{howard1980formulae}.

Mathematics expressed in formal systems is called \emph{formal mathematics}. It is expressive: Almost all mathematics can be expressed by first-order logic with ZFC set theory~\cite{fraenkel1973foundations}. At the same time, it enforces formal constraints: Formulas must conform to grammar rules, and their manipulation must conform to inference rules that capture valid reasoning. This is similar to how board games like chess and Go are played within predetermined rules and moves. The success of AI on board games~\cite{hsu2002behind,silver2016mastering} suggests that a similar approach could be applied to formal mathematics, even though mathematics, with an infinite number of configurations and moves, can be much more challenging than Go.

Specifically, formal systems can be useful environments for AI to learn mathematics. A formal environment can guarantee the soundness of reasoning, provide automatic feedback, and check if the goal has been achieved. This is crucial to addressing the two challenges faced by the informal approach: data scarcity and evaluation. Automatic feedback can serve as learning signals and alleviate the need for human-annotated training data. Rigorous proof verification allows us to evaluate the model's reasoning without worrying about hallucination.

\smallsec{Proof Assistants and Lean}
A concrete type of formal systems is called \emph{proof assistants}, also known as interactive theorem provers. These are software tools that enable humans to write formal proofs about mathematics or verified software. Common examples of proof assistants include Coq~\cite{barras1997coq}, Isabelle~\cite{nipkow2002isabelle}, and Lean~\cite{de2015lean,moura2021lean}. They have different logical foundations but share similarities from a user's perspective, regardless of whether the ``user'' is human or AI. For simplicity, we will frequently use Lean as an example to explain key concepts in formal mathematical reasoning, though many ideas can be applied to other proof assistants or formal systems in general.

\begin{figure}[ht]
  \centering
  \includegraphics[width=1.0\linewidth]{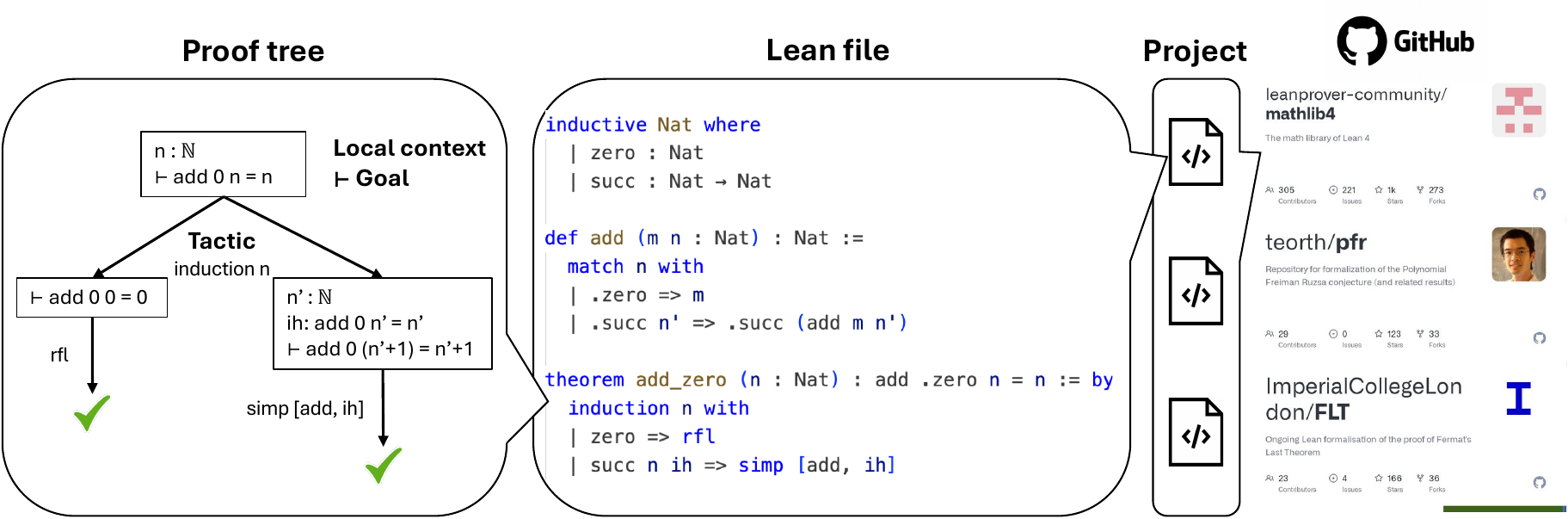}
  \caption{Formalizing mathematics using the Lean proof assistant~\cite{de2015lean,moura2021lean}.}
  \label{fig:lean}
\end{figure}

Fig.~\ref{fig:lean} demonstrates how Lean is used to formalize mathematics. At its core, Lean is a functional programming language with dependent types~\cite{christiansen2023functional}, making it suitable for writing not only conventional programs but also mathematical definitions, theorems, and proofs. Fig.~\ref{fig:lean} (\emph{Middle}) is an example Lean file. First, it defines natural numbers (\texttt{Nat}) as either zero or the successor of another natural number. Then, it defines addition between two natural numbers (\texttt{add}) as a recursive function. Finally, it states and proves the theorem \texttt{add\_zero} (\texttt{$\forall n \in \mathbb{N}, 0 + n = n$}). Lean can automatically check if the proof is correct with respect to the theorem statement. Technically, due to the Curry-Howard correspondence, statements in Lean are types, and proofs are expressions. As a result, proof checking is essentially type checking, i.e., verifying if an expression has the specified type.

Let's take a closer look at the proof of \texttt{add\_zero} (Fig.~\ref{fig:lean} \emph{Left}). In Lean, theorem proving is a backward and interactive process. It begins with the theorem statement as the initial goal, and the user enters a proof step, known as a "tactic". When executed by Lean, the tactic transforms the current goal into a list of sub-goals that are hopefully simpler. The user then inspects the new goals and enters new tactics, repeating this process until there are no goals left. This process implicitly defines a proof tree whose nodes are goals and edges are tactics (Fig.~\ref{fig:lean} \emph{Left}). The user plays a key role in theorem proving. While proof assistants like Lean were designed with human users in mind, in formal mathematical reasoning, the user can also be AI or humans in collaboration with AI (Sec.~\ref{subsec:tools}).

Formalizing mathematics using Lean is similar to software development, as shown in Fig.~\ref{fig:lean} (\emph{Right}). Lean files are organized into larger code units such as libraries and projects, which can be open-sourced on GitHub and reused by other projects. For example, the formalization of cutting-edge mathematical research like \href{https://github.com/teorth/pfr}{\texttt{pfr}} (the Polynomial Freiman-Ruzsa conjecture~\cite{gowers2023conjecturemarton}) often builds upon the basic concepts formalized in \href{https://github.com/leanprover-community/mathlib4}{\texttt{mathlib}}~\cite{mathlib}, Lean's general-purpose mathematical library. \texttt{Mathlib} currently contains 82,847 definitions and 161,483 theorems, covering a wide range of topics including analysis, algebra, and geometry. It is one of the largest monolithic repositories of formal mathematics and includes a substantial portion of postgraduate-level mathematics.

\smallsec{AI Meets Formal Mathematics}
Integrating AI with proof assistants such as Lean can benefit both AI researchers and the proof assistant community (mathematicians, computer scientists, and formal verification engineers). On the one hand, proof assistants provide data and environments for developing AI for formal mathematical reasoning. On the other hand, AI can enhance the user experience of proof assistants by automating simple proofs and suggesting useful lemmas.

\begin{figure}[ht]
  \centering
  \includegraphics[width=1.0\linewidth]{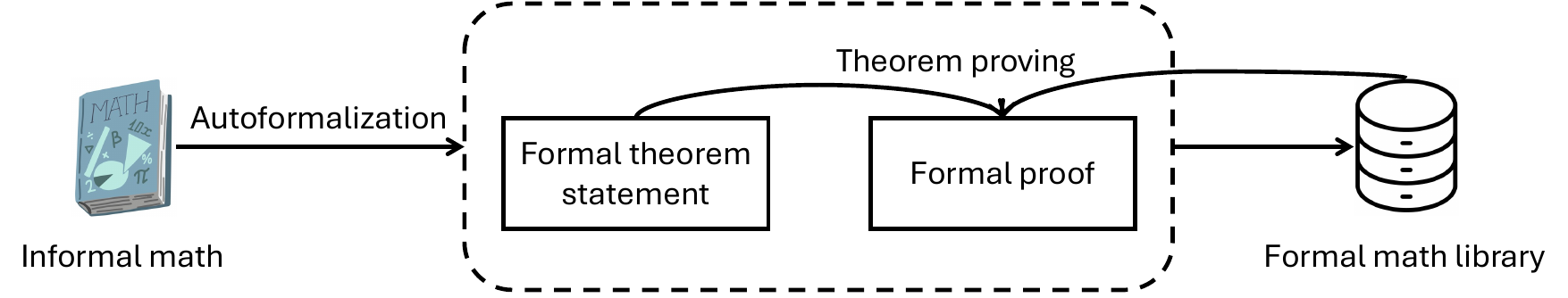}
  \caption{Common tasks using AI for formal mathematical reasoning in proof assistants.}
  \label{fig:overall}
\end{figure}

Fig.~\ref{fig:overall} includes common tasks at the intersection of AI and formal mathematics in proof assistants. Given informal mathematics written by humans (e.g., textbooks and papers), \emph{autoformalization} automatically translates it into formal theorems and proofs. Given theorem statements, \emph{theorem proving} aims to generate formal proofs. In addition to the statement, a theorem prover may have access to a large library of existing definitions and lemmas, such as \texttt{mathlib}, and can select useful definitions and lemmas from the library. Furthermore, AI for autoformalization and theorem proving can lead to new theorems and/or proofs that can enrich the library and bootstrap its own capability.

\begin{figure}[ht]
  \centering
  \includegraphics[width=1.0\linewidth]{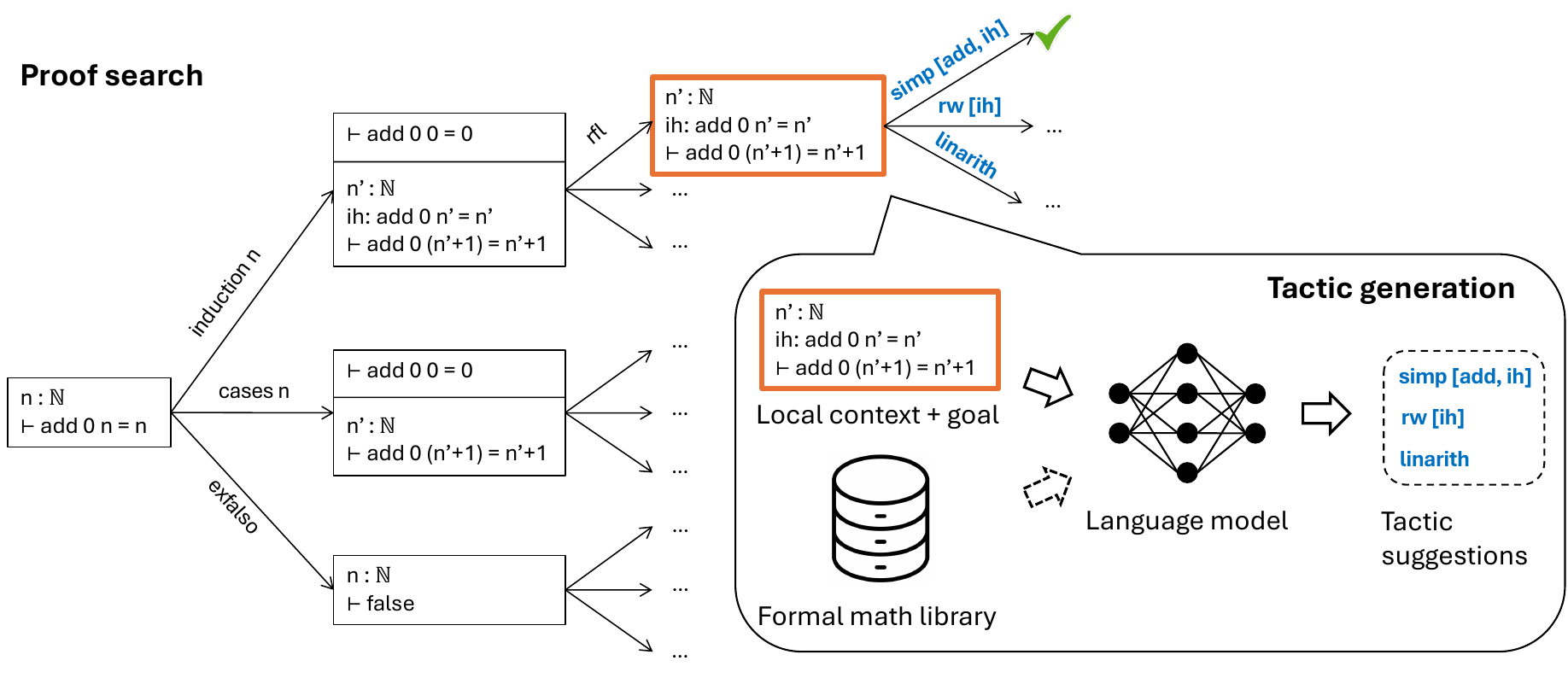}
  \caption{A neural theorem prover that combines tactic generation and proof search. This architecture is adopted by the majority of existing methods, with only a handful of exceptions~\cite{first2023baldur,xin2024deepseek}.}
  \label{fig:theorem_proving}
\end{figure}

Fig.~\ref{fig:theorem_proving} illustrates an architecture commonly adopted by recent neural theorem provers, consisting of two parts: tactic generation and proof search. Given the current proof goal (and optionally a library of definitions and lemmas), a neural network generates suggestions for the next tactic. The network is often trained on human-written proofs and can be further finetuned using reinforcement learning. The generated tactics are assembled into a complete proof by the proof search algorithm. It starts from the theorem statement as the root and grows a search tree iteratively by executing tactics to expand nodes, until a proof is found. The order of expansion is decided by the search algorithm, e.g., classical algorithms such as best-first search (BFS) and Monte Carlo tree search (MCTS).

\begin{figure}[ht]
  \centering
  \includegraphics[width=1.0\linewidth]{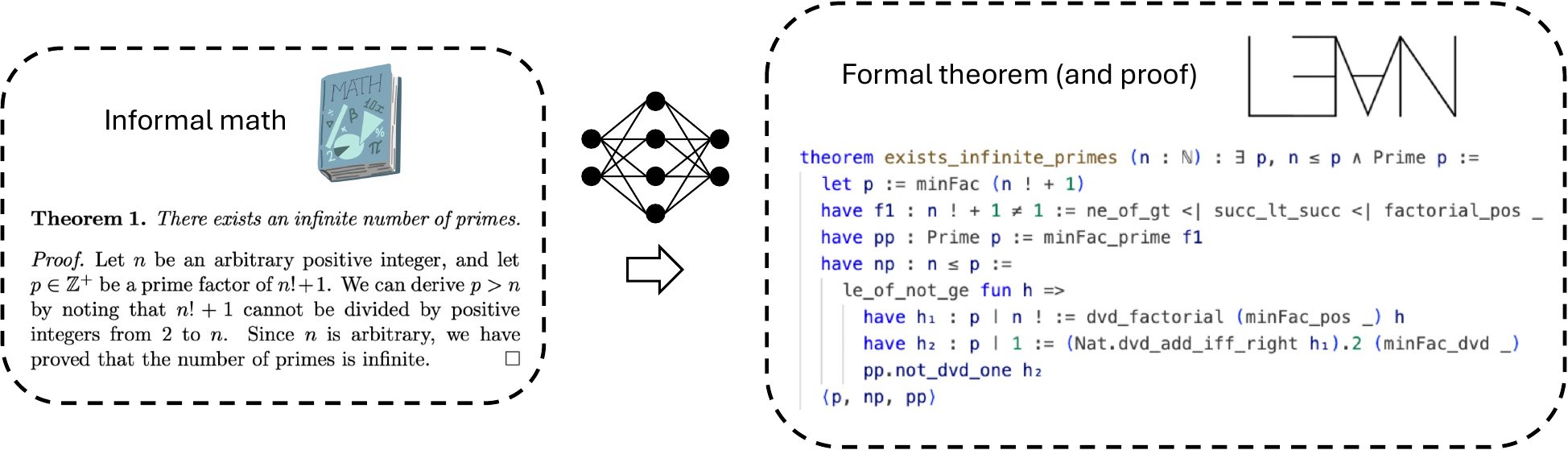}
  \caption{Autoformalization translates informal math to formal theorems and/or proofs automatically.}
  \label{fig:autoformalization}
\end{figure}

Fig.~\ref{fig:autoformalization} illustrates the task of autoformalization, i.e., translating from informal to formal. It takes informal mathematics (e.g., textbooks and papers) and translate it into theorem statements in a formal system such as Lean. In some settings, autoformalization also translates the proof~\cite{jiang2023draft,murphy2024leaneuclid}, which can be viewed as a form of theorem proving given informal proofs as hints.

The formal and informal approaches to reasoning should not be viewed as mutually exclusive, nor do we argue that formal reasoning should entirely supplant informal reasoning. Instead, these approaches can potentially complement each other to enable complex reasoning that is both general and rigorous. For example, autoformalization can be integrated with theorem proving to solve problems formulated in natural language~\cite{zhou2024don}. The model can generate reasoning steps in natural language while attempting to autoformalize part of the reasoning, which obtains feedback from the formal environment and helps filter out invalid reasoning. We refer to this integration of formal and informal reasoning as \emph{verified reasoning in natural language}, which is discussed in detail in Sec.~\ref{subsec:verified-natural-language} and \ref{sec:milestone-natural-language}.

\subsection{Other Directions in AI for Mathematics}
\label{subsec:others}

While we have highlighted the distinction between informal and formal approaches, AI4Math is a broad and open-ended research field that does not fit neatly into this dichotomy. For example, in addition to generating solutions or proofs, neural networks can also be used to approximate mathematical functions. This includes simple functions such as greatest common divisor~\cite{charton2024learning}, eigenvalues~\cite{charton2021linear}, and modular arithmetic~\cite{lauter2024machine}. These functions can be computed by well-known algorithms; however, approximating them through neural networks offers valuable insights into the model's capabilities and mechanistic interpretability. Furthermore, neural networks can approximate more complex functions for which we do not have efficient algorithms, with applications in cryptography~\cite{wenger2022salsa}, theoretical physics~\cite{cai2024transforming}, control theory~\cite{alfarano2024global}, and partial differential equations~\cite{raissi2017physics}.

Collaborative efforts between AI researchers and mathematicians have led to notable progress on open problems in mathematics. For example, FunSearch~\cite{romera2024mathematical} uncovered new solutions to the cap set problem, a long-standing open problem in additive combinatorics~\cite{tao2006additive}. It represents solutions as programs and leverages LLMs to generate new programs from existing ones, using an automated evaluator to assess the quality of solutions. \citet{gukov2021learning} use reinforcement learning (RL) to generate transformations for knot simplification in topology, while \citet{wagner2021constructions} uses RL to find counterexamples to open conjectures in graph theory. PatternBoost~\cite{charton2024patternboost} finds mathematical constructions by combining Transformers with classical local search algorithms, demonstrating effectiveness on open problems in graph theory and combinatorics~\cite{berczi2024note}. These works share similarities with formal mathematical reasoning: They also represent mathematical objects as symbols and manipulate them according to well-defined inference rules. However, they are tailored to specific subdomains rather than relying on general-purpose mathematical languages like Lean. Incorporating domain-specific mathematical insights can be valuable, as further discussed in Sec.~\ref{subsec:algorithms}.

\section{Recent Progress in AI for Formal Mathematical Reasoning}
\label{sec:progress}

AI has made substantial progress in formal mathematical reasoning. First, we discuss the progress in two key tasks: autoformalization and theorem proving. Then, we sample two adjacent areas---natural language and code generation---that benefit from verifiable reasoning enabled by the formal approach.

\subsection{Autoformalization}
\label{sec:autoformalization}

Pinpointing the origin of the phrase ``autoformalization'' is challenging, but the pioneers of automated theorem proving in the 1950s and 60s clearly conceived the idea:
\begin{displayquote}
\emph{The original aim of the writer was to take mathematical textbooks such as Landau on the number system, Hardy-Wright on number theory, Hardy on the calculus, Veblen-Young on projective geometry, the volumes by Bourbaki, as outlines and make the machine formalize all the proofs (fill in the gaps).} \hfill --- Hao Wang~\citep{wang1960toward}
\end{displayquote}
Nevertheless, it was not until the rise of interactive theorem proving (e.g., Automath~\cite{de1983automath}) that people began to seriously consider automating the labor-intensive formalization process~\cite{qed_manifesto}.

\smallsec{Rule-Based Autoformalization} Mohan Ganesalingam~\cite{ganesalingam2013language} explored a linguistic foundation for mathematical texts, introducing a theory of types to eliminate ambiguities in both words and symbols. To address the flexibility and complexity of natural languages, many systems have adopted controlled natural language---a restricted subset of natural language governed by formal grammar---to allow users to express mathematical proofs in a way that is both natural and formal. Examples of such systems include Mizar~\cite{grabowski2010mizar}, NaProChe~\cite{kuhlwein2009naproche}, ForTheL~\cite{vershinin2000forthel}, MathNat~\cite{humayoun2010mathnat}, and Verbose Lean~\cite{massot2024teaching}. Strictly speaking, these systems may not qualify as autoformalization since they do not directly handle natural language in its entirety. Simultaneously, a high-level grammar-writing tool called the Grammatical Framework (GF)~\cite{ranta2004grammatical} has been gaining attention. This tool allows for the flexible development of customized grammars to parse mathematical texts directly. GF-based systems include GLIF~\cite{schaefer2020glif}, a framework for symbolic natural language understanding and processing, and GFLean~\cite{pathak2024gflean}, an autoformalization framework that translates natural language statements into Lean’s formal language.

\smallsec{Neural and LLM-based Autoformalization}
\citet{kaliszyk2015learning, kaliszyk2017system} conducted early experiments using machine learning to parse informal mathematical texts, followed by \citet{wang2018first,wang2020exploration}, who applied neural machine translation techniques to convert informal mathematical statements into the Mizar language. Unlike rule-based methods, machine learning approaches are more flexible and can capture edge cases in natural language that experts might miss when creating rules. Although \citet{wang2020exploration} explored unsupervised machine translation~\cite{lample2018unsupervised}, most early machine learning methods for autoformalization relied heavily on an aligned corpus of informal and formal statements, which is difficult to obtain on a large scale.

LLMs like GPT-4~\cite{openai2023gpt4} represent a new paradigm of machine learning. These autoregressive models are pretrained on vast amounts of Internet data and can be quickly adapted to various downstream tasks through a few demonstrations, without requiring parameter updates---a capability known as in-context learning. Wu~et~at.~\cite{wu2022autoformalization} showed that, with fewer than five expert-crafted examples, LLMs can translate between formal and informal mathematical statements to some extent. This finding is promising, as it suggests that we may not need to collect a large aligned corpus of informal-formal statements---an almost impossible task given the variety of formal systems---to achieve autoformalization. Another notable observation in \citet{wu2022autoformalization} is that (auto-)informalization is generally easier than (auto-)formalization. With the same model, about 30\% accuracy was achieved in formalizing competition-level math statements, while over 70\% accuracy was achieved in informalizing even more challenging undergraduate-level math problems. This insight prompted follow-up research utilizing LLMs-based back-translation, where synthetic aligned corpora were generated by auto-informalizing existing formal statements~\cite{jiang2023multilingual, azerbayev2023proofnet}. Finetuning a smaller model on this synthetic data led to notable improvements in autoformalization performance. Building on a synthetic corpus, \citet{lu2024process} further incorporated additional signals from the formal environment to develop a process-driven autoformalizer.

LLMs have also significantly advanced the translation of natural languages into formal \emph{domain-specific languages} (DSLs) like SQL~\cite{li2024dawn} and linear temporal logic~\cite{pan2023data,mavrogiannis2024cook2ltl}. In this paper, we primarily focus on the more expressive formal languages used in foundational proof assistants. These languages are capable of accommodating both statements and proofs of modern mathematics, but come with the challenge of non-static vocabularies (definitions and proof tactics can evolve or expand over time). Nevertheless, autoformalization and NL-to-DSL translation are closely related, sharing techniques such as self-consistency and self-correction.

\smallsec{Application of Autoformalization}
Autoformalization serves as a bridge between informal and formal mathematical knowledge, resulting in three immediate applications: (1) data argumentation for training neural proving agents (via autoformalizing informal theorem statements)~\cite{wu2022autoformalization,xin2024deepseek}, (2) guiding formal theorem proving via informal proofs~\cite{jiang2023draft, murphy2024leaneuclid}, and (3) validating informal reasoning~\cite{zhou2024don,olausson2023linc}. The first two applications are closely related to neural theorem proving, which will be discussed in detail in the next section (Sec.~\ref{subsec:current-theorem-proving}). The third application will be examined in Sec.~\ref{subsec:verified-natural-language}.

\subsection{Neural Theorem Proving}
\label{subsec:current-theorem-proving}

Proving theorems in any sufficiently expressive formal system is undecidable~\cite{godel1931formal}. Thus, theorem proving inevitably requires heuristics. Deep learning has been widely used for \emph{learning} heuristics to find proofs in formal systems. 
Holophrasm~\cite{whalen2016holophrasm} was the first system to demonstrate the feasibility of training deep neural networks to guide proof search. Holophrasm used the Metamath formal language~\cite{megill2019metamath} as its logical backbone, training gated recurrent unit (GRU) networks~\cite{chung2014empirical} on human-written formal proofs to serve as policy and value networks in Monte Carlo Tree Search (MCTS)---the same search algorithm used in AlphaGo~\cite{silver2016mastering}. This training paradigm was expanded in GPT-f~\cite{polu2020generative}, which trained a single Transformer model for predicting proof steps in Metamath. GPT-f enjoyed substantial gains from pretraining on informal math-related data (arXiv Math, Math StackExchange, Github). Subsequent approaches have trained richer architectures such as retrieval-augmented Transformers~\cite{yang2023leandojo} and also exploited zero-shot prompting of general-purpose LLMs~\cite{thakur2024language,first2023baldur}.
We now highlight several prominent ideas in this literature. 

\smallsec{Expert Iteration}
Since a formal theorem proving environment can guarantee the validity of proofs, whenever a model finds a proof of a new theorem, that proof can be used as new training data. In the context of theorem proving, expert iteration consists of alternating between (a) attempting a set of unsolved problems, and (b) finetuning the model using new training data produced during the first phase. This has been shown to lead to improvements in formal theorem proving~\cite{polu2023formal,lample2022hypertree}, including in the recent work of AlphaProof~\cite{alphaproof}. However, gains tend to diminish after a few iterations. It is still an open problem to obtain continuous improvements, leveraging the potentially unlimited feedback that a formal verifier can provide.

\smallsec{Learning from Mistakes} 
A key benefit of formal proof environments is that they can provide error messages when a proof step fails. COPRA~\cite{thakur2024language}, an approach that repeatedly asks a frontier LLM to predict proof steps from within a search loop, uses such error messages as part of the LLM's prompts. COPRA is also equipped with a memory that stores a subset of the incorrect predictions that it made while proving a particular theorem, and this memory is included in the LLM's prompt. Because of frontier LLMs' ability to learn in context, these strategies reduce the odds of similar mistakes being repeated again and again. However, not all mistakes result in immediate failures; many lead to distractions or unproductive paths without making meaningful progress toward the proof goal. Identifying and learning from such mistakes remains a challenge.

\smallsec{Informal Proof Sketches} Neural theorem proving in \emph{formal} languages has also benefited from \emph{informal} proofs, generated in natural language. Notably, Draft, Sketch and Prove (DSP)~\cite{jiang2023draft} proposed to first have the LLM generate a ``proof sketch'', in natural language, to then attempt to formalize this sketch in Isabelle. Lean-STaR~\cite{lin2024lean} proposed to interleave formal and informal reasoning steps in theorem proving in Lean. COPRA, mentioned above, also takes in informal proofs as an optional input and uses them as part of zero-shot proof-step-prediction queries to a general-purpose LLM.

\smallsec{Library Learning} Human mathematicians can leverage a continuously growing library of mathematical results to find new ones. Each new theorem potentially enters this library for reuse later in higher-level work. Building towards this behavior, research in AI for formal mathematics has started to explore the idea of \emph{library learning}, where not only the search heuristics (neural policy and value functions) improve with more data, but also the symbolic library that is available. Library learning first gained popularity in the context of program synthesis, with systems such as DreamCoder~\cite{ellis2023dreamcoder} that are capable of discovering increasingly higher-level abstractions from its solutions to previous tasks. For theorem proving, LEGO-Prover~\cite{xin2023lego} demonstrated this idea in Isabelle, by proposing potentially reusable lemmas to aid in the proof of a theorem at hand. Besides new theorems, researchers have also explored learning \emph{tactics}---procedural proof-generation strategies that shorten otherwise lengthy low-level proofs, often tailored to a particular mathematical domain. Tactic induction has been demonstrated to work in simpler formalisms~\cite{poesia2023peano,schulz2000learning}, but has not yet been developed for full-fledged environments such as the tactic languages in Lean or Coq.

\smallsec{Premise Selection and Retrieval}
A large, changing library of theorems poses challenges for training neural theorem proving models, since the prover should not be limited solely by the lemmas and definitions it was given at training time. One architecture that can accommodate a changing library is retrieval-augmented generation (RAG). In RAG, before attempting to \emph{generate} (e.g., a proof for a target theorem), we first \emph{retrieve} potentially useful pieces of data from a database (e.g., lemmas from a mathematical library), putting these in the context that is given to the LLM. Retrieving lemmas that are likely to be used to prove a target theorem is a problem known as \emph{premise selection}~\cite{kuhlwein2012overview}, which has been extensively studied both in learning-guided ~\cite{alama2014premise,irving2016deepmath,piotrowski2020stateful} and symbolic~\cite{kuhlwein2013mash,meng2009lightweight,jiang2022thor} theorem provers. Even before neural networks were directly applied to generating proofs, they had been shown to be effective as premise selection models in earlier works like DeepMath~\cite{irving2016deepmath}.

ReProver, the architecture introduced in LeanDojo~\cite{yang2023leandojo}, applied retrieval for neural theorem proving, where it first retrieves lemmas from a mathematical library. The COPRA~\cite{thakur2024language} approach also uses retrieved lemmas as part of LLM queries. One desideratum in such retrieval-based approaches, originally identified in LeanDojo, is success on a train/test split named ``new premises'', where each theorem seen at test time requires at least one premise that has not been seen during training. This setup is closer to real use cases for a theorem prover, where human mathematicians might be formalizing a new domain, frequently building on recently proved lemmas that did not exist in the original training data.

\subsection{Verified Reasoning in Natural Language} \label{subsec:verified-natural-language}

Many reasoning problems expressed in unstructured natural language are difficult to formalize completely. In such cases, it is desirable to still have \emph{some} form of verification for natural language reasoning. Several works have proposed to ``verify'' natural language reasoning via trained, few-shot prompted, or symbolic verifiers. For example, the paper introduced the GSM8K dataset~\cite{cobbe2021training} of grade-school mathematical word problems proposed finetuning GPT-3 to verify a partial solution. Combining GPT-3 as a solution generator with the finetuned verifier produced substantial accuracy gains. On this line of work, OpenAI later released PRM800K, a large-scale dataset of human-written feedback on step-by-step mathematical solutions, which has been used to explore training verifiers~\cite{lightman2024let}. For logical reasoning in natural language, NLProofS~\cite{yang2022nlproofs} proposed a step-wise verifier, trained to evaluate whether a given conclusion is entailed by a set of premises, using it to guide a step-by-step prover. Step-by-step verifiers can also be obtained by prompting an LLM in a context dedicated to verification, as done in Natural Programs~\cite{ling2024deductive}. In all of these works, while the verifier cannot formally guarantee the validity of the reasoning, it nonetheless provided a boost in overall performance and faithfulness of the responses.

As opposed to verifying directly in natural language, another line of work has explored using LLMs to first formalize a problem given in natural language, then leverage a symbolic solver to find solutions. In this thread, SatLM~\cite{ye2023satlm} and LINC~\cite{olausson2023linc} have both leveraged SAT/SMT solvers as a logical backend for reasoning problems, with the LLM only being responsible for parsing the original problem into an appropriate formal counterpart. In both methods, however, the system does not provide a step-by-step solution, since all the reasoning happens inside the solver, in a form that is challenging to translate back to natural language. LogicGuide~\cite{poesia2024certified} proposed to use a formal system to constrain the step-by-step deductions from the LLM, producing chain-of-thought reasoning that alternates between formal and natural language. In all of these systems, while the formal reasoning is guaranteed to be sound, it is still difficult to assess whether the natural language problem has been properly formalized. This typically stems from the fact that natural language allows for ambiguity and reliance on common sense or general world knowledge, whereas the formal problem must be fully unambiguous and all of the assumptions must be fully written down, which is typically challenging.

\subsection{Formal System Verification and Verified Generation}

The formal verification of software \cite{hoare1969axiomatic,leroy2009formal,ringer2019qed} and hardware systems \cite{seligman2023formal,goel2022microprocessor} has long been among the foremost applications of formal mathematics. 
In this area, one first specifies the correctness and security requirements of a system as formal assertions. Next, theorem proving and model-checking techniques are used to prove that the system satisfies its requirements or, alternatively, to find bugs.

Specifically, deductive theorem proving 
has been applied to a range of critical systems, including microprocessor designs \cite{brock1996acl2}, file systems~\cite{chen2015using,chajed2022verifying}, OS kernels~\cite{klein2009sel4,gu2016certikos}, cryptographic implementations~\cite{bhargavan2013implementing,barthe2011computer}, compilers~\cite{leroy2016compcert}, distributed systems~\cite{sharma2023grove}, and network infrastructure~\cite{bhargavan2017everest,zheng2023automated}. However, writing formal specifications and proofs that are necessary for verification requires substantial manual effort. For instance, verifying the seL4 OS kernel involved more than 20 person-years of intricate engineering work~\cite{klein2014proof}.

AI for formal mathematical reasoning offers a promising way to automate many tedious aspects of theorem proving applied to system verification, drastically reducing its costs. Advances in neural theorem proving, as detailed in Sec.~\ref{subsec:current-theorem-proving}, can be effectively harnessed in software and hardware verification efforts that use formal mathematics, facilitating the generation of initial proofs~\cite{sanchez2020generating,thakur2024language,sanchez2025qedcartographer} and the refinement of existing proofs~\cite{first2023baldur,lu2024proof}. Moreover, LLMs are useful for assisting with various SMT-based verification tasks, including inferring necessary loop invariants~\cite{pei2023can,wu2024lemur,loughridge2024dafnybench,si2020code2inv}, generating helpful assertions~\cite{mugnier2024laurel}, and translating natural language to formal specifications~\cite{cosler2023nl2spec}. 

A closely related challenge is employing AI to simultaneously generate code with formal proofs of correctness and security. For example, LLMs have recently made remarkable progress in programming tasks~\cite{chen2021evaluating}. However, LLM-generated code can be buggy and insecure~\cite{pearce2022asleep,pan2024lost}, 
and some recent research finds LLM-generated code to exhibit more vulnerabilities than human-written code~\cite{perry2023users}. Coupling generation with formal verification is a natural way to prevent such failures. 

One possibility here is to first develop a formally verified program (or design) in a framework like Coq and Lean, with AI assistance, and then to translate the developed artifact into a more efficient lower-level implementation using standard compilers. This approach establishes a direct arc between theorem proving and generation. Another possibility is to incorporate LLM-based code and proof generation into a high-level verification-friendly language like Dafny \cite{misu2024towards} or Verus~\cite{lattuada2023verus}. 

The design of formal specifications is a particularly challenging aspect of formal methods. However, the autoformalization techniques mentioned in Sec.~\ref{sec:autoformalization} can help generate formal specifications from natural language or code. There are also settings such as transpilation \cite{bhatia2024verified} in which specifications come for ``free''. 
In transpilation, one starts with code for a system in a source language and uses AI to generate code in a different target language. The two systems are required to be equivalent; thus, the source-language code forms a complete specification for the target-language code.

\section{Open Challenges and Future Directions}
\label{sec:roadmap}

Formal mathematical reasoning presents a wealth of challenging problems for AI. Here, we explore several open challenges and promising directions, including data and algorithms for formal mathematical reasoning, AI tools for assisting human mathematicians and proof engineers, as well as integrating AI and formal methods to generate verifiable code. While the discussion inevitably reflects our own perspectives and preferences, we hope it will provide inspiration and a roadmap for the broader community to advance in this field.

\subsection{Data}
\label{subsec:data}

\begin{tcolorbox}[enhanced, colback=black!5!white, colframe=black!75!white, title=How to overcome the scarcity of formal data?]
\begin{itemize}
\itemindent=-35pt
    \item Autoformalizing informal math from textbooks, research papers, and lecture notes.
    \item Generating synthetic conjectures and proofs from mathematical axioms.
    \item Knowledge transfer from different proof frameworks and data-rich modalities such as code.
\end{itemize}
\end{tcolorbox}

A key driver of the performance improvements in LLMs has been captured in empirical scaling laws: LLM performance tends to broadly and consistently improve when we grow the model size and data size together~\cite{hoffmann2022training,zhang2024scaling}. However, this pace of improvement due to scale alone faces significant challenges when applied to formal mathematics, due to the scarcity of human-created formal proof data.
For instance, the Proof Pile dataset~\cite{azerbayevllemma}, which aggregated proofs from six different formal languages (Lean, Coq, Isabelle, HOL Light, Mizar~\cite{jakubuuv2023mizar}, and Metamath~\cite{yu2024metamath}), collected only 500MB of formal proofs. This is orders of magnitude smaller than relevant datasets for training LLMs in other domains, such as Python code available on GitHub. The data scarcity issue is even more pronounced in research-level mathematics, where even informal data is limited. At the research frontier, data will arguably always be scarce; by the time we have abundant data on a particular domain, there might not be much novel research left to be done there.

Researchers have been exploring a few different strategies to overcome data scarcity. The first is autoformalization: attempting to automatically formalize informal mathematical texts. We have substantially more informal math data available on the web in the form of textbooks, research papers, lecture notes, and other resources, far exceeding the current formal math libraries. One of the hopes in the field is to create a positive reinforcement loop when attempting to automatically formalize these sources. Since formal proofs can be mechanically checked, if a system successfully translates even a small subset of the available informal math data, it can learn from those translations for training in an expert iteration loop, potentially covering an increasingly larger set with each iteration (Sec.~\ref{subsec:current-theorem-proving}).

The second approach relies on synthetic data generation using a formal system. AlphaGeometry~\cite{alphageometry} recently took this approach, completely eschewing pretraining on human-written problems and instead relying solely on synthetically generated geometry problems and solutions. This strategy leverages the fact that mathematical axioms contain, in principle, \emph{infinite} potential data, since they entail all of the provable facts in the domain. If it can be generalized, a significant benefit of this approach would be its applicability to completely new domains of mathematics, where even informal data (such as research papers) might be scarce. By generating synthetic data, AI systems can potentially explore and learn from the vast space of possible mathematical problems and solutions, at a scale that can drastically surpass the pace of human-generated training data.

Autoformalization and synthetic data generation were combined in AlphaProof~\cite{alphaproof}, which autoformalized one million IMO-like informal problems into one hundred million \emph{formal theorems}, whereas \emph{proofs} were synthetically generated in an expert iteration loop. It remains an open question to generalize this approach beyond domains where a large number of human-written problems are available, as will be the case in research mathematics. For those domains, we will likely also depend on \emph{conjecturing} new, unseen statements~\cite{poesia2024learning}. 

A third approach is the use of multilingual data. Different formal proof frameworks tend to have different distributions of proof data; for example, formal software verification efforts have historically used Coq and Isabelle more than Lean, while recent efforts to formalize research mathematics are largely Lean-based. Building AI systems that can interact with different proof environments \cite{thakur2024language} is one way to get the best of different proof frameworks. An alternative path is to translate formal theorems and proofs across frameworks, but this direction remains relatively unexplored as of now. 

Another promising strategy to enhance AI’s capabilities in formal mathematical reasoning is transferring knowledge from different modalities. Specifically, the code modality is closely related to formal mathematics as both require symbolic reasoning. This similarity has been exploited to improve AI’s skills in mathematics during both inference~\cite{gao2023pal} and training phases~\cite{guo2024deepseek,dubey2024llama}. Prior research has shown that multi-lingual code models often outperform mono-lingual models when provided with equivalent training resources~\cite{athiwaratkun2022multi} and knowledge of high-resource languages can be transferred to low-resource ones using program translation~\cite{cassano2023knowledge}. This raises an interesting research question: how can we leverage datasets of data-rich programming languages such as Python and C/C++ to enhance reasoning in formal mathematical languages? Moreover, considering that current AI models for formal mathematics typically focus on single languages~\cite{huang2019gamepad,yang2023leandojo,lin2024fvel}, there is a promising opportunity to develop a multi-lingual model (e.g., combining Lean, Coq, and Isabelle), potentially boosting performance across all these languages.

\subsection{Algorithms}
\label{subsec:algorithms}

\begin{tcolorbox}[enhanced, colback=black!5!white, colframe=black!75!white, title=How to scale up autoformalization?]
\begin{itemize}
\itemindent=-35pt
    \item Automatic metrics for evaluating autoformliazed statements.
    \item Breaking the autoformalization process into small steps (like in chain-of-thought).
    \item Autoformalizing with more interaction with the formal system.
\end{itemize}
\end{tcolorbox}

\smallsec{Autoformalization at Scale} 
As a bridge between informal and formal mathematics, it should be emphasized that autoformalization is a task beyond pattern matching. For example, given an informal statement we might formalize it in multiple ways that are syntactically different but logically equivalent. Alternatively, two syntactically similar formal statements can certainly bear opposite logical meanings. As \citet{jiang2023multilingual} observed, neither classic automatic machine translation metric like BLEU~\cite{papineni2002bleu} nor compiling success rate really correlates with human evaluation, but relying on human evaluation is obviously not scalable. Without an automatic metric to evaluate autoformalized statements, it would be challenging to build a robust statement formalization pipelines. Nevertheless, wrongly autoformalized statements can still be utilized: In AlphaProof~\cite{alphaproof}, DeepSeek-Prover~\cite{xin2024deepseek}, and Lean Workbook~\cite{ying2024lean}, 
agents attempt to simultaneously prove and disprove autoformalized statements, and either will leave a trace to further reinforce the neural proof agents. Still, we want to have an automatic metric more aligned with human judgment so that the quality of autoformalized statements can be truly evaluated at scale. Possible ideas include checking logical equivalence via automated theorem provers~\cite{li2024autoformalize,murphy2024leaneuclid}.

Depending on the level of mathematical abstraction, autoformalization can range from a relatively straightforward translation task (e.g., formalizing `$1+\cdots+n = n(n+1)/2$') to a hard reasoning task (e.g., formalizing `every finite group of odd order is solvable') that requires retrieving existing definitions or even inventing new ones. For a hard reasoning task, it is natural to break down the autoformalization process into smaller steps, like in chain-of-thought prompting. For example, \citet{wu2022autoformalization} observed that a common source of errors in autoformalizing statements is the mismatch between informal and formal definitions, which might be alleviated by retrieving definitions before attempting to formalize the statement. When autoformalizing proofs, a natural approach is to convert a large piece of informal proof into a formal sketch before attempting to fill in the details. An alternative approach, used in Lean-STaR~\cite{lin2024lean}, is to sample one natural-language step each time and use it to synthesize the next proof tactic. We envision that through smaller steps and process supervision~\cite{lightman2024let}, the performance of existing autoformalization models could be further improved~\cite{lu2024process}.

Finally, given that formalization is an interactive process for human experts, often involving extensive trial and error with the proof assistant, we believe that a good autoformalizer should inherently support such interactivity. Formalizing statements may require introducing new definitions and data types (e.g., in the 2024 IMO P5), while autoformalizing proofs could benefit from a recursive approach to address gaps in informal proofs. Both aspects call for autoformalization to be more interactive~\cite{szegedy2020promising}.

\begin{tcolorbox}[enhanced, colback=black!5!white, colframe=black!75!white, title=How to improve the model architecture for mathematical reasoning?]
\begin{itemize}
\itemindent=-35pt
    \item Multi-step reasoning, long contexts, abstractions, and hierarchical planning.
    \item Controlled studies on synthetic benchmarks for diagnosing reasoning failures.
    \item Scaffolding the model with inference-time techniques such as retrieval and search.
\end{itemize}
\end{tcolorbox}

\smallsec{Models for Mathematical Reasoning}
Machine learning models for mathematics should possess two key capabilities: First, the model should be capable of memorization, allowing it to store mathematical knowledge, such as facts, definitions, and existing theorems. Second, the model should be able to reason effectively about its knowledge,  which necessitates multi-step reasoning, handling long contexts, learning abstractions, and hierarchical planning. In recent years, Transformer-based Language models~\cite{vaswani2017attention} have become the leading architecture for AI mathematical reasoning~\cite{lewkowycz2022solving,polu2020generative,wu2022autoformalization}. Transformers excel at memorization: They are the most scalable architectures to date for Internet-scale pretraining~\cite{hoffmann2022training,zhang2024scaling}, during which their model parameters are adapted to internalize mathematical knowledge in the pretraining data. Although a precise understanding of how Transformers extract and memorize knowledge is still nascent~\cite{allen2024physics}, their effectiveness is widely recognized in practice~\cite{petroni2019language}.

However, whether Transformers can reason logically is an open question. On the one hand, they have demonstrated exceptional performance on numerous reasoning benchmarks~\cite{hendrycks2021measuring,suzgun2023challenging}. On the other hand, they still exhibit reasoning flaws, even in simple settings~\cite{mirzadeh2024gsm}. For example, \citet{wang2024grokked} show that Transformers learn to reason only through grokking, which occurs when a model is trained far beyond the point of overfitting---a scenario not applicable to pretrained LLMs. \citet{zhang2023paradox} find that Transformers fail to learn true reasoning that is generalizable across different data distributions. Furthermore, LLMs have been found to fall short on planning~\cite{valmeekam2023planning,valmeekam2024planbench,xie2024travelplanner}.

Formal mathematics provides a valuable domain for understanding and improving Transformers' capabilities in reasoning and planning, as well as for developing alternative architectures~\cite{peng2023rwkv,gu2023mamba,ye2024diffusionthoughts,du2024learning,ho2024proof}. When experimenting with new architectures, we often face the dilemma that these models underperform standard LLMs because they do not readily benefit from large-scale pretraining, whereas limiting to pretrained models would restrict the model choice. Formal mathematical reasoning may help address this dilemma through synthetic benchmarks~\cite{wu2021int}. These benchmarks can be procedurally generated based on the formal system's inference rules, and they can have adjustable knobs for controlling the difficulty levels or testing a specific capability, such as generalization to longer reasoning chains~\cite{anil2022exploring}. Additionally, performing well on these simple benchmarks typically requires only small models that can be trained within days using a few GPUs. This setup enables controlled scientific experiments, which help diagnose the model's failures and discover insights that can potentially be scaled up in follow-up studies.

Besides designing an architecture that inherently excels at reasoning, another highly effective approach is to scaffold the model with external techniques like chain-of-thought~\cite{wei2022chain}, retrieval~\cite{yang2023leandojo}, and search~\cite{lample2022hypertree}. Next, we will discuss some of these topics in detail.

\begin{tcolorbox}[enhanced, colback=black!5!white, colframe=black!75!white, title=How to search for proofs effectively?]
\begin{itemize}
\itemindent=-35pt
    \item Scaling up the search to leverage more test-time compute.
    \item Systematic evaluation of models, search algorithms, and hyperparameters.
    \item Value models for assessing and prioritizing proof goals.
\end{itemize}
\end{tcolorbox}

\smallsec{Proof Search and Test-Time Compute}
\begin{displayquote}
\emph{Search and learning are the two most important classes of techniques for utilizing massive amounts of computation in AI research.} \hfill --- Richard Sutton~\citep{sutton2019bitter}
\end{displayquote}

Proof search is fundamental to many formal reasoning systems. In particular, most existing neural theorem provers (Fig.~\ref{fig:theorem_proving}) generate proofs by combining tactic generation with search algorithms such as breadth-first search~\cite{bansal2019holist}, best-first search~\cite{yang2023leandojo}, or Monte Carlo Tree Search (MCTS)~\cite{lample2022hypertree}. Scaling up the proof search to leverage an enormous amount of test-time compute has been crucial to the success of AlphaGeometry~\cite{alphageometry} and AlphaProof~\cite{alphaproof} on IMO problems. Furthermore, despite originating from formal reasoning~\cite{newel1976computer}, proof search has recently gained prominence in natural language reasoning~\cite{lightman2024let,yang2022nlproofs,zhang2024rest,xie2024monte}. There, LLMs' reasoning is scored by a verifier model (also referred to as ``process-supervised reward model'') and is searched to arrive at a final solution, much like searching for a proof. Scaling test-time computation during search has led to promising results~\cite{snell2024scaling} and is likely a key component of OpenAI o1~\cite{o1}.\footnote{This is largely speculative, as there is limited public information about the internals of o1.}

Despite the importance of proof search, many myths and trade-offs surrounding it remain unexplored. First, \emph{is proof search truly necessary?} Baldur~\cite{first2023baldur} and DeepSeek-Prover~\cite{xin2024deepseek} use LLMs to generate whole proofs directly, without search. This approach offers substantially lower latency, making it attractive in interactive applications such as proof completion in code editors. A widely held belief supporting proof search is that decomposing proofs into individual steps can improve data efficiency and generalization. While plausible, we are unaware of empirical evidence from a systematic comparison between search and whole-proof generation. Second, \emph{should we use small models or big models in proof search}~\cite{wu2024empirical}\emph{?} Given a fixed compute budget, smaller models allow exploring more steps. For instance, Graph2Tac~\cite{blaauwbroekgraph2tac} suggests that simple models like k-nearest neighbors can perform competitively with Transformers. In addition, \emph{how do different search algorithms compare (e.g., MCTS vs. best-first search) ? What is the effect of decoding algorithms (e.g., sampling vs. beam search) and hyperparameters (e.g., temperature)? How to search for proofs efficiently with access to high-end GPUs (typical in AI research) vs. consumer CPUs (typical among Lean users)?}

To resolve these myths, it is necessary to systematically evaluate existing theorem proving methods, as this will provide clarity and guide the development of future provers. However, such an evaluation is currently lacking and would require substantial effort. Conducting a fair and unified evaluation of theorem provers presents significant challenges. It is unclear how to compare provers targeting different proof assistants. While cross-system benchmarks like MiniF2F~\cite{zheng2022minif2f} and PutnamBench~\cite{tsoukalas2024putnambench} have formalized problems across multiple proof assistants such as Isabelle and Lean, this alone does not imply the results are comparable. The difficulty of proving a theorem can vary widely between proof assistants due to the varying levels of proof automation. Even within a single proof assistant, a theorem prover's performance is multifaceted and depends on resource constraints (e.g., hardware and time limits), making it difficult to consolidate performance into a single metric. A comprehensive evaluation that carefully addresses these challenges would be immensely valuable to the community.

Despite the challenges in evaluation, researchers have been exploring various directions to improve proof search. Effective search requires prioritizing the most promising goals for further exploration. Mathematicians often rely on intuition to gauge a goal's promise, and the popular MCTS algorithm can be viewed as its Monte Carlo estimation. One fruitful direction is developing value models for assessing proof goals, through finetuning~\cite{lample2022hypertree,wu2024internlmstepprover} or by prompting instruction-following LLMs as demonstrated in Tree of Thoughts~\cite{yao2023tree}. However, assessing the promise of proof goals remains a challenging task. Minor changes in the goal can impact its provability. Furthermore, while positive examples (promising goals) can be extracted from existing proofs, negative examples are much harder to obtain. One approach is to use the current model to generate negative examples during proof search and try to bootstrap the model carefully. Alternatively, when working in a narrow domain, we can leverage domain-specific knowledge to evaluate proof goals (more on domain-specific provers later).

\begin{CJK*}{UTF8}{gbsn}
Proof search is important, but it alone does not solve theorem proving. Unlike Go, a fundamental challenge in theorem proving is a discrete, infinite action space whose structure is not fully understood. This unbounded nature makes it difficult for models to generate effective actions through supervised learning or exploration via reinforcement learning. Proof search cannot succeed if the model cannot produce high-quality actions in the first place. Many believe that mathematics requires creativity, and in the context of theorem proving, creativity can manifest as actions exceeding the current model's capabilities, akin to the ``divine move (神之一手)''---a legendary concept in Go. We would not expect to find them if the action space were an infinite, unstructured list. Fortunately, mathematics is highly structured, making it possible---though still challenging---to find the divine moves~\cite{gowers2022how}. In the remainder of this subsection, we will explore several ways to leverage structures in mathematics, including hierarchies, abstractions, external knowledge, and domain-specific knowledge.
\end{CJK*}

\begin{tcolorbox}[enhanced, colback=black!5!white, colframe=black!75!white, title=How to exploit hierarchies in theorem proving?]
\begin{itemize}
\itemindent=-35pt
    \item Decomposing large, high-level proof goals into smaller goals progressively.
\end{itemize}
\end{tcolorbox}

\smallsec{Exploiting Hierarchies in Theorem Proving}
Theorem proving can benefit from exploiting the natural hierarchies that organize mathematical results: Big theorems follow from smaller lemmas, and even those lemmas can be thought of as progressively accomplishing smaller sub-goals during the proof, until each goal is small enough to be ``obvious'' in informal proofs, or closed in a single step in a proof assistant. Several existing theorem proving systems attempted to exploit this hierarchy. For example, Draft, Sketch, and Prove (DSP)~\cite{jiang2023draft} used an informal proof (written by LLMs or humans) to obtain a formal ``proof sketch''---a skeleton of the formal proof with ``holes'', i.e., open goals left unproven, which yields a hierarchical structure for the formal proof. However, even a single open goal in the proof sketch might require significant effort to prove. POETRY~\cite{wang2024proving} proposed to recursively decompose proof goals using an LLM. It verifies that each of the intermediate decompositions is valid: When a larger proof goal is decomposed into several smaller goals, it must be provable, assuming the smaller goals can be proved. On another use of hierarchy, LEGO-prover~\cite{xin2023lego} proposes separate, potentially helpful lemmas, that it tries to prove first when it fails to prove a given theorem directly. While these works have begun to exploit the potential of hierarchical decomposition, it is still a significant challenge to decompose realistic high-level goals (or sometimes even individual informal proof steps) with the current capabilities of LLMs. Ideally, we would like humans to be able to provide high-level targets, and let AI do the work of progressively closing the gaps between what is currently known (e.g., the current proof state, or the existing library) and what it needs to achieve.

\begin{tcolorbox}[enhanced, colback=black!5!white, colframe=black!75!white, title=How to learn mathematical abstractions?]
\begin{itemize}
\itemindent=-35pt
    \item Learning to construct new definitions, lemmas, and tactics in full-fledged proof assistants.
\end{itemize}
\end{tcolorbox}

\smallsec{Learning Mathematical Abstractions}
While learning mathematics, humans can learn progressively more sophisticated mathematical abstractions. We start learning and operating on natural numbers by counting; years later, those operations show up in solving equations, but don't require as much attention anymore. Later on, even solving entire systems of equations becomes the simplest step in the context of harder problems. One research challenge is how to allow machines to progressively construct these abstractions. In interactive theorem provers, these abstractions are encapsulated in new definitions, lemmas, and tactics---these encode proof strategies that are helpful in a particular domain or kind of proof goal. In principle, these forms of abstraction aren't essential for formally representing mathematics, since what they do can always be repeated inline in a given context. Yet, progressively developing new abstractions is central to the human practice of mathematics.

Most theorem proving systems focus on taking in a set of definitions, lemmas, and tactics, leveraging those to prove new theorems. However, several recent lines of research have proposed methods for learning abstractions. The interactive theorem proving community has developed several methods for \emph{lemma synthesis}~\cite{yang2019lemma}, typically with the goal of helping a user prove a particular theorem interactively~\cite{sivaraman2022data}. In AI for mathematics, LEGO-Prover recently used LLMs to propose and prove new lemmas that also get added to its library, helping it prove further theorems. Lemma \emph{mining} from existing proof corpora has also been explored, such as in HOL Light~\cite{kaliszyk2013lemma} and Metamath~\cite{zhourefactor}: these lemmas, not explicitly factored out by humans, are still useful for automation. On learning \emph{tactics}, Peano~\cite{poesia2023peano} and LEMMA~\cite{li2022lemma} have proposed to learn simple proof strategies from an agent's own solutions to past mathematical problems, in a bootstrapping fashion. These so far have been demonstrated only in simpler formal systems, and it is still an open challenge to synthesize entirely new tactics in full-fledged formal theorem proving languages.

\begin{tcolorbox}[enhanced, colback=black!5!white, colframe=black!75!white, title=How to utilize existing mathematical knowledge?]
\begin{itemize}
\itemindent=-35pt
    \item Tailored retrievers for formal mathematical reasoning.
    \item Handling dynamically growing knowledge bases.
\end{itemize}
\end{tcolorbox}

\smallsec{Incorporating Information from Knowledge Bases} 
Another direction is to explicitly incorporate knowledge from databases of pre-existing lemmas and definitions. To some extent, LLMs used in math reasoning tasks are repositories of mathematical knowledge. However, some of the knowledge relevant to proofs may not be represented in the LLM's pretraining data; even if this knowledge were present, it may not be easy to retrieve. Therefore, the use of an explicit retrieval mechanism can help.

Among existing methods, ReProver~\cite{yang2023leandojo} and COPRA~\cite{thakur2024language} use retrieval mechanisms and achieve nontrivial performance gains from their use. These approaches use standard retrieval mechanisms, such as BM25~\cite{robertson2009probabilistic} and Dense Passage Retrieval~\cite{karpukhin2020dense}. It is possible that retrieval mechanisms more tailored to formal mathematics would perform better. For example, one can imagine developing retrieval methods based on structured, neurosymbolic embeddings that use vector representations of math facts while also allowing symbolic methods to filter out irrelevant facts.

Another angle is to consider scenarios in which the external knowledge base grows over time. For example, one can imagine a mathematical reasoning system that decomposes high-level proof objectives into subgoals, caches a subset of these subgoals as modules, and appropriate modules for use in subsequent (or concurrent) proof efforts. Deciding which subgoals are ``interesting'' enough to be modularized in this way is an interesting challenge.

\begin{tcolorbox}[enhanced, colback=black!5!white, colframe=black!75!white, title=How to reconcile the specialist and generalist approaches?]
\begin{itemize}
\itemindent=-35pt
    \item Generalist methods for identifying cross-domain connections.
    \item Specialists for effectiveness in individual domains and collaboration with mathematicians.
    \item Combining generalists and specialists, e.g., by equipping LLMs with domain-specific tools.
\end{itemize}
\end{tcolorbox}

\smallsec{Generalist vs. Specialist}
Mathematics encompasses a wide range of subdomains, and so far, our discussions have largely remained domain-agnostic. In principle, most mathematical domains can be formalized in proof assistants such as Lean, enabling LLMs to perform tasks such as theorem proving and autoformalization. LLMs only need to process the data as plain text, without accounting for the specifics of each domain. This ``generalist'' approach has clear merits: it is generally applicable and facilitates knowledge sharing across domains. Modern mathematics is too vast for any individual to master everything. LLMs, however, with massive training data, can easily surpass human mathematicians in terms of breadth of knowledge. Numerous historical mathematical breakthroughs resulted from uncovering connections between seemingly unrelated subjects. For example, Wiles's proof of Fermat's Last Theorem~\cite{wiles1995modular} emerged from identifying connections between elliptic curves and modular forms. In the long term, it is desirable to have LLM-powered generalist AI to augment human mathematicians in identifying such cross-domain connections. 

However, each mathematical domain possesses its own idiosyncrasies and unique techniques. The generalist approach risks missing the opportunity to exploit these domain-specific insights. Many AI4Math systems are specialists in a particular domain. A notable example is AlphaGeometry~\cite{alphageometry}, which specializes in proving Euclidean geometry theorems in math olympiads. It consists of three key components: (1) an algorithm that generates synthetic theorems and proofs; (2) a symbolic reasoning engine for deducing basic geometric properties; and (3) a Transformer model for introducing auxiliary constructions (points, lines, or circles not present in the original problem). Each component is designed specifically for 2D Euclidean geometry and leverages domain knowledge to achieve efficiency. While some high-level ideas in AlphaGeometry could potentially be applied to other domains, adapting the entire system would require substantial redesign.

Specialists like AlphaGeometry are highly effective in their domains, often solving problems that are currently challenging to address in a fully general manner. These systems leverage domain knowledge in various ways, e.g., for proof automation, evaluation, finding counterexamples, or numerical computation~\cite{murphy2024leaneuclid,wagner2021constructions,wei2024proving,anonymous2025proving}. Building specialists for other mathematical domains will continue to be a fruitful research direction, particularly for collaborating with mathematicians who want to use AI for their domain of interest (Sec.~\ref{subsec:others}). In addition, narrow domains offer a rich set of controllable tasks and environments~\cite{wu2021int} for investigating the reasoning capabilities of AI models.

The Bitter Lesson~\cite{sutton2019bitter} stresses the great power of general-purpose methods, and AI researchers always strive for generality. Even when designing specialists, we value insights that have the potential to generalize. Despite this, we anticipate that AI math specialists will coexist with generalists in the foreseeable future. An exciting avenue lies in combining the two, such as by equipping a general LLM with domain-specific algorithms as tools. Ultimately, AI might become powerful enough to learn and exploit the unique characteristics of each domain in a universal way.

\subsection{Tools for Assisting Human Mathematicians}
\label{subsec:tools}

\begin{tcolorbox}[enhanced, colback=black!5!white, colframe=black!75!white, title=How can AI better assist humans in formal mathematics?]
\begin{itemize}
\itemindent=-35pt
    \item Resources, incentives, and engineering efforts to improve usability and user-friendliness.
    \item Behavioral studies of how mathematicians work with formal tools.
    \item Tools that enable large, distributed collaboration. 
\end{itemize}
\end{tcolorbox}

While improving performance on standardized benchmarks is a means to coordinate efforts and measure progress in AI, many significant challenges only arise when trying to build tools that integrate with the workflow of mathematicians using proof assistants. Working mathematicians often report that the bottleneck for the adoption of tools, rather than technical features, is their \emph{usability} and \emph{user-friendliness}~\cite{drosser2024ai,ai4mathfund}. To productively integrate AI tools for assisting mathematicians, the field can drastically benefit from behavioral studies of mathematicians working with formal tools in an ecological fashion. We note that the Human-Computer Interaction and Programming Languages communities have both collaborated in such studies for users of regular programming languages, leading to insights into how human programmers learn and conceptualize their tools~\cite{crichton2021role}, their confusions and challenges, and ultimately how to improve these tools and the way we teach them~\cite{crichton2023grounded}. We envision that many methodologies from such work can be transported to proof assistants, ensuring that the tools we build serve the needs of human mathematicians.

Besides aiding the workflows of individuals, one key advantage of proof assistants that has been highlighted is that they enable large-scale collaboration among mathematicians~\cite{ringer2024proofs, drosser2024ai}. Tools such as Lean Blueprints~\cite{leanblueprint} allow large mathematical projects to be conceptually broken down into modular components, and each of those can be worked out independently. Trust in other's work, which traditionally has either required personal trust or understanding and checking all of their work, is now facilitated by a formal proof checker. These tools are likely to be just a first generation of what formal proof assistants can provide for collaboration in mathematics. In the future, collaboration might happen between both humans and computers in a distributed fashion~\cite{ringer2024proofs}; AI agents might make autonomous contributions to human projects, and vice-versa, mediated by formal tools that guarantee correctness. This vision can likely borrow from previous experiences in distributed, collaborative computing platforms from other fields, like Folding@Home~\cite{voelz2023folding}, where anyone can contribute computing power to protein folding. The same is possible for proving mathematical theorems. 

\subsection{Formal Verification and Verified Generation}
\begin{tcolorbox}[enhanced, colback=black!5!white, colframe=black!75!white, title=How can AI assist humans in developing correct and secure software?]
\begin{itemize}
\itemindent=-35pt
    \item Incorporating formal methods into AI-aided system design and implementation.
    \item Enhancing AI capabilities for formal software and hardware verification.
    \item Coupling AI-based generation and formal verification.
\end{itemize}
\end{tcolorbox}

In alignment with our stand for AI4Math, we envision a growing necessity for developing formal reasoning techniques for AI-based software and hardware generation. These techniques can provide developers with assurance on the correctness and security of generated artifacts---an indispensable step for deployment, which currently often requires significant manual effort. While syntactical correctness can be guaranteed by constrained decoding~\cite{beurer2024guiding,ugare2024improving}, ensuring other semantic properties, such as those validated by static analysis and compilers, remains an open challenge. Moreover, formal reasoning can assist programmers in understanding AI-generated code that they did not write themselves. Recent research demonstrates this through the live demonstration of runtime values~\cite{ferdowsi2024validating}. It is a promising direction to explore the use of other kinds of formal methods in this context.

Formal verification bears some resemblance to the research mathematics setting but also poses unique challenges. For example, a necessary but challenging step for formal verification is encoding the target system semantics and the correctness requirements in the proof assistant. This process is akin to formalization of informal mathematics~\cite{wu2022autoformalization}; however, while statements in mathematics research tend to assert properties of established mathematical objects, theorems in formal verification typically concern bespoke procedures and datatypes. Also, proofs in formal verification tend to be more repetitive and heavy on case-splits and inductive reasoning about recursive functions and datatypes. 
Finally, unlike statements in mathematics research, real-world software and hardware systems are characterized by large codebases and frequent changes. For instance, the verification of the seL4~\cite{klein2009sel4} operating system kernel consists of about 200,000 lines of specifications and proofs in Isabelle. Verification of these systems requires not only theorem proving but also rigorous management of specifications and proofs~\cite{ringer2019qed}. An exciting yet underexplored direction is leveraging AI’s strong capabilities in code and mathematics to enhance the entire process of proof engineering~\cite{ringer2019qed}.

It is natural to couple formal verification and AI-based generation into approaches that simultaneously generate code (or designs), formal specifications (i.e., pre-/post-conditions, loop invariants, and helper assertions), and proofs. Given these generated artifacts, a program verifier or a theorem prover can be called to check if the code is consistent with the specifications and proofs. This approach has been explored in recent research~\cite{sun2024clover,misu2024towards,yang2024autoverus,aggarwal2024alphaverus} and holds great potential in reducing verification efforts and enhancing software and hardware reliability. However, a key challenge is ensuring the trustworthiness of the generated specifications---that they are neither too strong nor too weak and accurately reflect developers’ intent.

\section{Milestones and Success Measures}
\label{sec:milestones}

Having outlined directions for advancing this field, a key question is: \emph{how can we effectively measure progress?} Inspired by the levels of automation for self-driving cars~\cite{sae2024taxonomy}, we propose a framework for categorizing AI's capabilities in formal mathematical reasoning. Focusing on individual areas such as theorem proving and autoformalization, we define various capability levels and review existing benchmarks for evaluation. Existing benchmarks fall short when it comes to evaluating higher-level capabilities, particularly in nascent areas like conjecturing. To address these gaps, we highlight the pressing need for new benchmarks and, in many cases, new evaluation methodologies.

\subsection{Theorem Proving}
\label{sec:milestone-proving}

\begin{table*}[ht]
  \small
  \centering
  \caption{Theorem proving: capability levels and benchmarks for evaluation.}
  \begin{tabular}{l p{6cm}  p{6cm}}
    \toprule
     Level & Capability & Evaluation and benchmarks \\
    \midrule
    0 & Checking formal proofs & Achieved by modern proof assistants\\
    \midrule
    1 & Assisting humans to develop proofs by suggesting definitions, lemmas, proof steps, etc. & Human-centered evaluation \\
    \midrule
    2 & Human-implemented tactics for automating domain-specific proof goals & Domain-specific benchmarks \\
    \midrule
    3 & Proving simple theorems automatically in a domain-general fashion & CoqGym~\cite{yang2019learning}, LeanDojo~\cite{yang2023leandojo}, MiniF2F~\cite{zheng2022minif2f}, PutnamBench~\cite{tsoukalas2024putnambench}, TPTP~\cite{sutcliffe1998tptp}, FIMO~\cite{liu2023fimo} \\
    \midrule
    4 & Contributing to formalization projects autonomously & New benchmarks of code and metadata from GitHub, similar to SWE-bench~\cite{jimenez2024swe} \\
    \midrule
    5 & Solving problems and discovering new math beyond the human level & New benchmarks and evaluation methodology for unknown territories \\
    \bottomrule
  \end{tabular}
  \label{table:theorem_proving_levels}
\end{table*}

Most of the current effort in AI for formal mathematics has been centered around \emph{automated theorem proving}. Here, given a theorem, we want AI to produce a proof. Formal systems, such as Lean, offer an immense advantage to pose this problem, since once a proof is found, it is guaranteed to be correct, even if it might be difficult to parse by humans. We now discuss milestones and benchmarks in AI for formal theorem proving (Table~\ref{table:theorem_proving_levels}).

\textbf{The most basic capability level (Level 0) is simply recognizing a correct formal proof.} It is already present in systems such as Lean~\cite{de2015lean,moura2021lean}, Coq~\cite{barras1997coq}, Isabelle~\cite{nipkow2002isabelle}, Agda~\cite{bove2009brief}, and many others, and can be obtained from a wide range of logical foundations. Though proof checking is essentially a solved problem for \emph{formal proofs}, it can still be incredibly challenging to fully formalize even existing, well-understood \emph{informal} proofs, to the level where they can be mechanically verified by a formal system. Human mathematicians, in contrast, are able to evaluate incomplete, ``hand-wavy'' arguments, often being able to find flaws and point out counterexamples even in the absence of a complete proof. Thus, even our proof checking systems can still be improved in this direction by requiring less effort from the user and putting more responsibility on the verifier. \emph{We currently do not have a good benchmark for the verification of incomplete, high-level proofs}. Since these proofs often have large gaps that are left unwritten, checking them requires being able to fill those gaps, which is generally equivalent to the full problem of producing proofs (which the next capability levels are about). For benchmarking progress in formalizing mathematics, one folkloric standard used to compare different proof assistants is Freek Wiedijk's list~\cite{formalizing100theorems} of one hundred theorems. As of today, all theorems in the list have already been formalized (by humans) in at least one proof assistant.

Moving beyond giving feedback on proof correctness, we can think of varying levels of systems that can help humans develop proofs. \textbf{Level 1 would be to suggest potentially useful pieces of data, without attempting to write proofs yet.} Here we include library search engines, such as Moogle~\cite{moogle}, Loogle~\cite{loogle}, and LeanSearch~\cite{leansearch}, which can find useful theorems or definitions, or ``copilots''~\cite{copilot,song2024towards} that can generate contextual completions interactively. One common use case is to search for relevant definitions from the current goal and see what is already known about those (e.g., see theorems involving the \texttt{sin} function). These tools can be highly helpful, though the main job of deciding what to search for and how to develop the proof is still a human responsibility.

\textbf{Capability Level 2 and above contain systems that generate proofs, either fully or partially. The most basic level of proof automation is (human-implemented) tactics:} domain-specific procedures that are capable of automating certain classes of proofs, often serving as simpler steps in larger proofs. An example would be a tactic such as \texttt{omega} in Lean and Coq, which can automatically solve a large class of equality and inequality proof goals, or hammer tactics~\cite{sledgehammer}, which outsource the current proof goal to an external (most commonly first-order) automated theorem prover. While limited in domain, these procedures can already lower the burden on human users to formalize results. Still, up to this level, no learning-based system is involved: this level of automation mostly reflects the human ingenuity required to engineer domain-specific methods for producing proofs.

\textbf{At Level 3, we include systems that can automatically prove theorems in a domain-general fashion, albeit still limited to simple theorems.} This encompasses most current neural theorem provers~\cite{yang2023leandojo,wei2024proving,wang2024proving}: current systems are typically trained on human-written proofs, and can generally be applied to any domain. Several of the most recent benchmarks in neural theorem proving target this level of capability: CoqGym~\cite{yang2019learning}, LeanDojo~\cite{yang2023leandojo}, MiniF2F~\cite{zheng2022minif2f}, PutnamBench~\cite{tsoukalas2024putnambench}, TPTP~\cite{sutcliffe1998tptp} and FIMO~\cite{liu2023fimo} are prominent examples of such evaluations. However, in practice, we are still limited to relatively simple proofs, typically not the most time-consuming ones in the context of a larger project. Moreover, current systems generally assume a static library of definitions and previous lemmas, and they target well-posed statements (in fact, in these evaluation benchmarks, we know all the statements to be true, since humans have already proved them). These systems can still be helpful by proving technical lemmas or closing gaps in larger proofs.

Level 3 is the last capability level covered by existing evaluations. One level above, we have all the activities that human mathematicians engage in while developing formalization projects. These go much beyond proving lemmas that are given to them, as evaluated in neural theorem proving benchmarks. In a new formalization, a large part of the effort lies in choosing how to formally describe the domain of interest in the first place: what definitions to use, and how to break down the main results of interest into sufficiently small lemmas.\footnote{Many definitions that are mathematically equivalent turn out to be very different in how easily formalizable they are in formal systems (e.g., formalizing real numbers as Cauchy sequences turns out to be easier in type theory than as Dedekind cuts, even though mathematically the theory of reals can be developed either way).} These activities happen at the formal and informal levels simultaneously: for instance, in Lean, mathematicians have been using Blueprints~\cite{leanblueprint} as a way to structure projects, divide up the work, and generally reason about how a large result might be broken down into manageable components. \textbf{A system of capability Level 4 should be able to plan and execute formalization projects autonomously, breaking down larger results, stating new definitions and lemmas, and potentially exploring different alternatives as the project develops.} This would already significantly accelerate progress in formal mathematics. For instance, we can expect recent papers to be formalized by AI at this stage. To evaluate Level 4 systems, it would be helpful to have new benchmarks constructed from the GitHub metadata, such as issues and commits, of real-world formalization projects. SWE-bench~\cite{jimenez2024swe}, a benchmark from software engineering, offers a model, but similar benchmarks for formal mathematics are yet to be created.

Finally, we might one day expect to have systems that go beyond what human mathematicians have been able to accomplish, solving existing open problems, or perhaps even formulating new interesting problems on their own. Currently, formal systems are typically used to formalize results that human mathematicians have first done informally. If one day AI becomes able to autonomously make mathematical discoveries, we can expect these discoveries to be made in a formal system. Otherwise, the cost of checking AI outputs will be too high, just as it currently is for verifying human mathematical proofs at the highest levels. \textbf{Being able to solve problems beyond human level would constitute capability Level 5.} While this seems categorically out of reach for any of the current AI systems, one fundamental challenge will be to be able to measure progress meaningfully towards this open-ended goal. A major difficulty with measuring progress towards mathematical discovery is that our current evaluations only test knowledge of current mathematics, while at this level we will want systems to be able to reason about new domains, which we ourselves might not know much about.

\subsection{Verified Reasoning in Natural Language} 
\label{sec:milestone-natural-language}

\begin{table*}[ht]
  \small
  \centering
  \caption{Verified reasoning in natural language: capability levels and benchmarks for evaluation.}
  \begin{tabular}{l p{6.5cm}  p{5.5cm}}
    \toprule
     Level & Capability & Evaluation and benchmarks \\
    \midrule
    0 & Stepwise natural language reasoning w/o verification & GSM8K~\cite{cobbe2021training}, AQuA~\cite{ling2017program} \\
    \midrule
    1 & Stepwise natural language reasoning with neural verification & MATH~\cite{hendrycks2021measuring}, Fallacies~\cite{hong2024closer}, ProcessBench~\cite{zheng2024processbench} \\
    \midrule
    2 & Tool-integrated reasoning using SymPy, NumPy, etc. & MATH~\cite{hendrycks2021measuring}, AIMO Progress Prize~\cite{aimo2024progress} \\
    \midrule
    3 & Reasoning seamlessly in natural language and formal systems such as Lean & New benchmarks for evaluating final answers and intermediate steps; problems difficult to formalize, e.g., IMO combinatorics \\
    \midrule
    4 & Complex mathematical reasoning and planning in real-world applications & Downstream applications such as travel planning and calendar scheduling \\
    \bottomrule
  \end{tabular}
  \label{table:verified_reasoning_levels}
\end{table*}

Theorem proving requires both the problem and the solution to be fully formalized, which can be overly rigid for many real-world applications. Even highly structured domains, such as IMO, have problems that are difficult to formalize (e.g., geometry and combinatorics problems), let alone everyday applications. \emph{How can we enable complex, rigorous reasoning without formalizing every aspect of a problem?} A promising direction is to use AI to reason seamlessly between formal systems and natural language. Such AI should be able to conduct logical reasoning, perform numerical calculations, and generate solutions in a way that is both rigorous and human-understandable. While the resulting reasoning chain may not constitute a formal proof, it could still include parts that can be verified semi-automatically, potentially under human oversight. We refer to this capability as verified reasoning in natural language and propose a framework for understanding its varying levels (Table~\ref{table:verified_reasoning_levels}).

\textbf{Level 0 involves generating step-by-step reasoning in natural language without verification.} The prevalent approach is to combine LLMs with chain-of-thought~\cite{wei2022chain}, but it frequently generates reasoning steps that are brittle, incorrect, or unfaithful~\cite{shi2023large,lanham2023measuring,ling2024deductive}. To address these limitations, \textbf{Level 1 capability introduces verification alongside generation, requiring AI to assess the correctness of reasoning steps.} This verification can be done by the same model that generates the reasoning~\cite{weng2023large} or by a different model~\cite{lightman2024let}. The result of verification can be used to improve reasoning. For example, we can generate many reasoning chains and use the verifier to select the best one~\cite{cobbe2021training}; the model can iteratively correct its own reasoning~\cite{pan2023automatically}; or it can search for solutions step by step, maximizing a correctness score produced by the verifier~\cite{yang2022nlproofs,zhang2024rest}.

Many existing benchmarks can be useful for evaluating Level 0 and Level 1, e.g., math word problems in GSM8K~\cite{cobbe2021training} and AQuA~\cite{ling2017program}, as well as commonsense reasoning and question answering benchmarks like CommonSenseQA~\cite{talmor2019commonsenseqa} and HotpotQA~\cite{yang2018hotpotqa}. Level 1 particularly requires benchmarks that are challenging for autoregressive generation. For instance, the MATH dataset~\cite{hendrycks2021measuring} includes competition-level math problems that require complex reasoning and planning, where the verification capability in Level 1 becomes essential~\cite{lightman2024let}. In addition to evaluating Level 1 indirectly through the final answer accuracy, it is valuable to directly measure whether the model can identify reasoning errors; Fallacies~\cite{hong2024closer} and ProcessBench~\cite{zheng2024processbench} are initial steps in that direction. 

One limitation with standard benchmarks like MATH is data contamination~\cite{dong2024generalization}, which may have partially contributed to LLMs' impressive, near-saturating performance. While data contamination is widely recognized, it is difficult to measure or fully eliminate its impact on evaluation outcomes. To mitigate data contamination, it is useful to dynamically generate benchmarks in a controllable manner. For example, the GSM-Symbolic~\cite{mirzadeh2024gsm} benchmark is generated by substituting various numbers into GSM8K-style templates. It effectively reveals reasoning flaws in current LLMs, calling into question whether they have achieved Level 0 and Level 1 robustly. Another potential solution to data contamination, adopted by FrontierMath~\cite{glazer2024frontier} and the AIMO Progress Prize~\cite{aimo2024}, is to keep the benchmark private and allows the model to access it during evaluation via a secure mechanism.

Moving to higher capability levels, we aim to make reasoning more rigorous and trustworthy. Relying solely on neural networks for verification becomes insufficient due to their brittleness and lack of interpretability. To address this limitation, the model can generate reasoning that can be \emph{partially} verified by external tools. These tools, built on well-established algorithms and thoroughly tested implementations, offer greater interpretability and trustworthiness than neural networks alone. \textbf{In Level 2, models can leverage external tools to perform computation that neural networks struggle to learn reliably}, such as numerical calculations with NumPy~\cite{harris2020array} and symbol manipulation with SymPy~\cite{meurer2017sympy}. Many recent math LLMs adopt this approach (called tool-integrated reasoning in Sec.~\ref{sec:mathllm}). Their performance can be evaluated on math problems requiring intricate computations, such as those in the MATH dataset or AIME. 

Tools like NumPy and SymPy are effective for computation but not suited for reasoning. \textbf{In Level 3, models should be able to use external tools to perform rigorous logical reasoning}. For instance, when generating reasoning chains, the model could interleave natural language with formal reasoning steps in Lean. Unlike theorem proving, where theorem statements are predefined, here the model dynamically generates the ``statements'' during inference. Furthermore, instead of formalizing the entire problem, the model can selectively determine which parts of reasoning to process using formal systems versus natural language, seamlessly integrating them to construct the solution. To our knowledge, no existing systems has achieved this kind of integration. Current approaches~\cite{ye2023satlm,pan2023logic,olausson2023linc,zhou2024don} attempt to autoformalize the entire problem and apply theorem proving techniques, which is challenging for problems that cannot be fully formalized.

With recent advances in tool-using LLM agents~\cite{gou2024tora}, the seamless integration of formal and informal reasoning may soon be within reach. However, new benchmarks are required to evaluate this capability. Unlike current benchmarks like MATH, these new benchmarks should evaluate not only the final answer but also the quality of reasoning that led to it. Additionally, they should include math problems that resist complete formalization. A promising candidate for such a benchmark could be IMO in the same format as human contestants,\footnote{Instead of the original IMO problems, AlphaGeometry~\cite{alphageometry} and AlphaProof~\cite{alphaproof} used problems manually formalized by human experts, which did not include combinatorics problems.} aligning with the evaluation protocol in the AIMO Prize~\cite{aimo2024}. During evaluation, human graders can bypass the formal reasoning steps and focus on checking the natural language reasoning and their interface with formal steps. Beyond correctness, we could also use verification effort as an additional metric: an ideal solution would be both correct and easily verifiable with minimal human effort.

Real-world applications that involve complex reasoning are often not purely mathematical problems, yet they contain significant mathematical components along with other components such as commonsense and human preferences. \textbf{Level 4 requires AI to recognize the mathematics in everyday tasks and apply rigorous reasoning.} For example, in scenarios like travel planning~\cite{xie2024travelplanner} or calendar scheduling~\cite{zheng2024natural}, AI can potentially formulate these tasks as constraint satisfaction problems and solve them using appropriate solvers such as mixed-integer linear programming (MILP) solvers. Achieving this capability would open up a wide range of new AI applications.

\subsection{Autoformalization}

\begin{table*}[ht]
  \small
  \centering
  \caption{Autoformalization: capability levels and benchmarks for evaluation.}
  \begin{tabular}{l p{5.8cm}  p{6.2cm}}
    \toprule
     Level & Capability & Evaluation and benchmarks \\
    \midrule
    0 & Representing knowledge in formal systems to support manual formalization & Achieved by modern proof assistants \\
    \midrule
    1 & Generating autoformalization candidates and collecting human feedback & Infrastructure to collect and store human feedback\\
    \midrule
    2 & Robust and faithful translations between informal and formal & ProofNet~\cite{azerbayev2023proofnet}, Herald~\cite{gao2024herald}; automatic evaluation of formal statements by equivalence checking \\
    \midrule
    3 & Inferring missing information and flagging situations when a gap cannot be filled & New benchmarks constructed from real-world formalization projects \\
    \cmidrule(r){1-2}
    4 & Self-correcting erroneous or inconsistent inputs by understanding human intentions & \\
    \cmidrule(r){1-2}
    5 & Proposing novel definitions that can reduce proof complexity\\
    \bottomrule
  \end{tabular}
  \label{table:autoformalization_levels}
\end{table*}

Autoformalization involves automatic translation between informal and formal representation of mathematical knowledge. \textbf{In Table~\ref{table:autoformalization_levels}, the most basic capability level (Level 0) is to store (i.e., encode and check) formal knowledge so that manual formalization is viable with enough human effort.} With the maturity of modern proof assistants, almost all mathematical proofs can be formally encoded and checked, thanks to the expressiveness of the underlying logic of those systems. Admittedly, some proofs involving diagrams or higher-dimensional structures could be challenging to formalize in existing systems, and may even require further study in alternative foundations, e.g., homotopy type theory~\cite{aczel2013homotopy}. We argue that most proofs with an axiomatic foundation can be tweaked into existing formal frameworks (the results may not look as natural as the informal proofs). With a formal encoding being viable, it is a matter of human efforts to transform the knowledge in different representations; therefore we consider this level to be mostly achieved.

\textbf{At Level 1, models should generate reasonable candidates for auto-(in)formalization, supported by an infrastructure to continuously gather and store human feedback, enabling a virtuous feedback loop to improve model performance over time.} Being exposed to both informal and formal knowledge, existing LLMs have already learned to align concepts and generate reasonable syntactical translation given formal/informal counterparts. However, what is mainly lacking at this level is a system to gather and store the feedback: humans might have interacted with the model to get the revised candidates and use them in a formal or informal setting (e.g., writing a paper), but neither the interaction process nor the end results (i.e., aligned informal-formal pairs) have been properly recorded. The Isabelle Parallel Corpus~\cite{bordg2022parallel} has been an early attempt to build a non-static parallel corpus. The Formal Abstracts project~\cite{fabstract}, initiated by Tom Hales, aimed to link standard mathematical publications with a Lean-based formulation that includes the main theorems but omits their proofs. By focusing solely on the statements, this project offered a potential framework for aligning contemporary mathematical statements with formalized counterparts. Unfortunately, the Formal Abstracts project did not gain traction for various reasons. As a less ambitious but more practical framework, Lean Blueprint~\cite{leanblueprint} focuses on individual formalization projects, allowing users to create human-readable ``blueprints'' that can later be linked to Lean formalizations. This approach has been successfully applied in high-profile projects such as the Liquid Tensor Experiment~\cite{liquid2022} and the formalization of the Polynomial Freiman-Ruzsa conjecture.\footnote{\url{https://teorth.github.io/pfr}} However, Lean Blueprint is used primarily as a one-off project planning tool, and informal proofs are not synchronized with their formal counterparts after the final formal proofs are integrated into Lean’s Mathlib. In short, current autoformalization pipelines are not yet mature enough for daily use. Even when autoformalization techniques are applied in a project (e.g., in Lean Blueprint), the results are recorded statically and in various formats (e.g., briefly mentioned in the comment sections of formal proofs), making them difficult to collect and prone to desynchronization as the formal statements evolve. A milestone we envision for this level is a centralized system that enables automatic translation between informal statements and, ideally, multiple formal representations. This system would allow users to submit revised results, and it would keep informal-formal pairs synchronized as the formal statements continue to develop.

Using the aligned corpus collected in Level 1, \textbf{AI models at Level 2 should be capable of performing robust and faithful translations between informal and formal statements, approaching human-level accuracy}. The main obstacle anticipated at this level is the automatic evaluation of translated (formal) statements. Early experiments have shown a misalignment between automatic metrics, such as parsing rate and BLEU scores, and human evaluations. Fortunately, recent work incorporating symbolic equivalence~\cite{murphy2024leaneuclid,li2024autoformalize, anonymous2024rethinking} may help establish a standard automatic metric that aligns better with human preferences for evaluating translated formal statements. Model performance at this level could be assessed using human-curated benchmarks, including challenges from the ICML 2024 Math-AI workshop~\cite{math_icml24_chanllenges,huang2024mustard}, ProofNet~\cite{azerbayev2023proofnet}, Herald~\cite{gao2024herald}, and Con-NF~\cite{anonymous2024rethinking}.

Levels 3, 4, and 5 go beyond pattern matching and focus more on reasoning. \textbf{Level 3 models should be capable of inferring missing information when autoformalizing statements and proofs, and can flag situations where an information gap cannot be filled.} When formalizing mathematics, we frequently deal with underspecified problems with missing or implicit assumptions in mathematical statements and hand-waived steps in proofs. Bridging these information gaps requires robust theorem proving and reasoning capabilities in the AI models: Proof gaps may be addressed through neural or symbolic theorem proving (Sec. \ref{sec:milestone-proving}), while missing assumptions can be resolved using abductive reasoning or counterexamples~\cite{bundy2005proof, blanchette2010nitpick}. The main challenge at this level is for the models to identify information gaps---such as assessing the likelihood that a statement is provable or can be adjusted---even when the reasoning model cannot bridge the gap immediately. This challenge is closely related to conjecturing (see Sec. \ref{milestone:conjecturing}) and has been examined briefly in some early explorations~\cite{bengio2024machine}. \textbf{At Level 4, AI models should be able to self-correct when they encounter erroneous or inconsistent inputs.} At this stage, the autoformalization model focuses more on capturing human intentions and may rely on its own self-consistency to eliminate errors. Advancements here will be closely linked to natural language reasoning (Sec.~\ref{sec:milestone-natural-language}). \textbf{Finally, at Level 5, AI models should be able to invent novel mathematical definitions that can potentially reduce proof complexity.} At this level, an AI model is closer to a ``theory builder'' that can reshape the proving process through better abstraction or concept formulation. For instance, filters (i.e., a set of sets satisfying certain properties) are rarely taught in standard math curriculum. However, as a convenient abstraction in mathematical analysis, they have become a widely adopted concept for formalizing limits in various proof systems, including Isabelle~\cite{holzl2013type}, Lean~\cite{mathlib}, HOL Light~\cite{harrison-euclidean}, and Coq~\cite{boldo2015coquelicot}. Automatically devising definitions like filters is what we hope an AI model can achieve at this stage.

\subsection{Conjecturing} \label{milestone:conjecturing}

\begin{table*}[ht]
  \small
  \centering
  \caption{Conjecturing: capability levels and benchmarks for evaluation.}
  \begin{tabular}{l p{7.3cm}  p{4.7cm}}
    \toprule
     Level & Capability & Evaluation and benchmarks \\
    \midrule
    0 & Generating conjectures useful for proving a target theorem & Existing theorem proving benchmarks \\
    \cmidrule(r){1-2}
    1 & Generating conjectures useful for proving theorems in a particular domain &  \\
    \midrule
    2 & Understanding interestingness computationally and generating interesting conjectures in new mathematical domains & New benchmarks and evaluation methodology for unknown territories  \\
    \bottomrule
  \end{tabular}
  \label{table:conjecturing_levels}
\end{table*}

While \emph{proving} theorems is a significant part of doing mathematics, a task that necessarily comes before proving a statement is coming up with the statement to prove in the first place. At that stage (before a proof), such a statement is a \emph{conjecture}, and we hope that AI agents might be able to formulate conjectures by themselves. Conjectures must not only be novel in a formal sense (e.g., different from all those made before), since it is possible to systematically generate new but uninteresting or unimportant conjectures. \emph{Interesting} conjectures, such as Fermat's last theorem or the Kepler conjecture~\cite{hales2008formal}, can often drive the work of mathematicians for centuries, as the pursuit of a proof or disproof unravels. But one fundamental challenge here is exactly in understanding what \emph{interesting} means in this context, and finding ways to evaluate it. Tackling this challenge would answer questions such as: \emph{How do we computationally define interestingness? What is the relationship between interestingness and usefulness? How does interestingness change over time, e.g., as special cases are proved or analogous results in other domains are disproved?}

\textbf{In Table~\ref{table:conjecturing_levels}, Level 0 would be to formulate conjectures in the context of a problem, or a particular target result.} These conjectures might be lemmas that, if proved, would help in proving a larger theorem of interest. This is related to how LEGO-Prover~\cite{xin2023lego} proposes lemmas for itself to attempt as helpers in the context of the current theorem. These simpler, targeted conjectures can be evaluated in the extent that they facilitate proving the theorems they were meant to support.

\textbf{More generally, at Level 1, we might expect a conjecturing agent to be able to formulate conjectures in a given \emph{domain}, without necessarily aiming at a particular theorem~\cite{urban2020first}.} While difficult to evaluate directly, these conjectures might still be evaluated in (a) how often the system can prove those conjectures and (b) whether the proofs of those conjectures, when found, serve as useful training data for proving other theorems in that domain. Minimo~\cite{poesia2024learning} is a system that implements this conjecturing-proving loop, showing that such conjectures can be generated only from the axioms of a given arbitrary domain, their difficulty can be targeted by a language model attempting to generate increasingly harder but provable conjectures, and that training on them improves the agent at unseen theorems. This paradigm has the potential to allow us to train systems for domains that have little or no human data available, which will necessarily be the case in new mathematical domains.  However, this so far has only been shown to be practical in relatively small domains (on the order of a few dozens of axioms). It is still an open challenge to do this at the level of large libraries, such as Lean's mathlib, and to maintain steady progress after many iterations of conjecturing (Minimo was tested with up to 5, and starts to saturate).

Going even beyond that frontier, \textbf{future AI systems might be able to formulate conjectures that go beyond current domains of mathematics.} Some expect that a major milestone for future AI systems will be to both \emph{conjecture and prove} a high-level, \emph{interesting} mathematical result. While that goes far beyond current systems, such an achievement would mark an era where AIs work side-by-side with human mathematicians in expanding our collective knowledge of mathematics.

\subsection{Formal Verification and Verified Generation}

\begin{table*}[ht]
  \small
  \centering
  \caption{Formal verification and verified generation: capability levels and benchmarks for evaluation.}
  \begin{tabular}{l p{5.7cm}  p{6.3cm}}
    \toprule
     Level & Capability & Evaluation and benchmarks \\
    \midrule
    0 & Code generation without verification & HumanEval~\cite{chen2021evaluating}, MBPP~\cite{austin2021program}, VerilogEval \cite{liu2023verilogeval} \\
    \midrule
    1 & Verifying simple properties of small programs and designs & miniCodeProps~\cite{lohn2024minicodeprops}; HumanEval/MBPP-Verus~\cite{aggarwal2024alphaverus}; new benchmarks for generating formal specifications and verifiable code \\
    \midrule
    2 & Verifying and synthesizing entire projects with complex functional and security properties & Selene~\cite{zhang2024selene} \\
    \midrule
    3 & Proof and system maintenance & Benchmarks constructed from the change history (GitHub metadata) of verified systems \\
    \cmidrule(r){1-2}
    4 & Helping users generate, explain, and debug formal specifications &  \\
    \bottomrule
  \end{tabular}
  \label{table:codegen_levels}
\end{table*}

As mentioned previously, the challenges in applying AI to formal verification and verified system generation are subtly different from those in the research mathematics setting.  It is natural to define a ladder of capabilities for this context based on the sophistication of the  generation and verification tasks (Table~\ref{table:codegen_levels}). \textbf{In capability Level 1, AI can handle small-scale verification tasks, which involve verifying small blocks of code, or small designs, against relatively simple properties, and synthesize small pieces of verified code.} This stage is critical as it sets the foundation for understanding how AI can be scaled up to real-world system verification efforts. Several existing benchmarks have targeted verification at this capability level~\cite{loughridge2024dafnybench,lohn2024minicodeprops,aggarwal2024alphaverus}. For instance, miniCodeProps~\cite{lohn2024minicodeprops} considers code properties that are viewed as a ``minimum level of competency" for automated neural theorem provers. However, GPT-4o can only prove around 35\% of the properties. 
Therefore, there is significant room for improvement in AI's ability to handle even these fundamental tasks. To the best of our knowledge, there is no existing high-quality benchmark for generating verifiable code together with the formal specification even at Level 0. 

As we move to larger-scale systems, the complexity of both the code and the specifications increases significantly. \textbf{In Level 2, AI should provide assistance in verifying and synthesizing entire projects and addressing complex properties.} Examples of such properties include preventing memory safety issues~\cite{calcagno2011infer}, enforcing access control~\cite{  klein2009sel4}, and proving the equivalence of programs~\cite{churchill2019semantic}. These efforts can improve the automation of various security-related tasks, such as C-to-Rust translation~\cite{yang2024vert} and reverse engineering~\cite{dasgupta2020scalable}. Achieving this level of verification involves decomposing large systems into smaller, verifiable components, a task that is currently performed by humans \cite{gu2016certikos}. Advanced AI techniques are required to tackle this challenge, such as agentic approaches~\cite{xie2024travelplanner,thakur2024language} that involve planning and problem-solving to navigate the intricate dependencies and interactions within large codebases and designs. Benchmarks for this level should incorporate project-level context, similar to repository-level code generation~\cite{zhang2023repocoder}. This can be achieved by repurposing existing verified systems~\cite{leroy2016compcert,gu2016certikos} to create tasks for AI. A recent benchmark called Selene explores this direction using the extensive codebases of seL4~\cite{zhang2024selene}.

System designs and implementations constantly evolve, and so must their proofs to ensure that the desired properties remain verified. \textbf{At Level 3, AI systems are expected to go beyond generation to proof and system maintenance.} When developers update the system or proof engineers decide to refactor proofs~\cite{ringer2021proof}, AI at this capability level should provide assistance, reducing the manual efforts needed for verification even further. To effectively evaluate AI tools at this capability level, benchmarks can be constructed by leveraging the change history of verified systems~\cite{reichel2023proof}. These benchmarks should capture a variety of scenarios, including minor bug fixes, major feature additions, and comprehensive refactoring efforts. The AI must demonstrate proficiency not only in generating proofs at a repository-level context~\cite{hu2024minictx} but also in reasoning about code and proof changes~\cite{first2023baldur}.

With Levels 1--3, AI systems possess the capabilities to generate and manage proofs, assuming the specifications expressing the properties that the generated artifacts must satisfy are provided. However, writing specifications is a significant challenge for formal verification, as it requires abstracting and converting user requirements into formal specification languages. \textbf{In capability Level 4, AI systems make another leap by aiding users in deriving formal specifications, including specification generation, explanation, and debugging.} To benchmark specification assistance, we can again leverage verified codebases and designs, but instead of generating proofs and code given the specifications, we treat the code and proofs as ground truth and use them to evaluate specifications produced by the model. AI systems need strong natural language understanding to interpret user requirements and translate them into formal specifications. They should also have interactive capabilities to engage with users, offering suggestions and clarifications. Moreover, they should be able to validate specifications against known best practices and standards, ensuring they are robust and comprehensive.

\section{Conclusion and Discussion}

In this position paper, we advocated for \emph{formal mathematical reasoning} as a new frontier in AI. By grounding reasoning in formal systems like Lean, this approach enables the training and evaluation of AI models whose reasoning can be rigorously verified, which holds the potential to significantly advance fields such as mathematics and software verification, as well as applications that require complex and rigorous reasoning. The advent of large language models has created opportunities for formal mathematical reasoning in AI, marking an inflection point for the field. For key tasks such as theorem proving and autoformalization, we discussed recent advancements, future directions, and milestones for measuring progress. AI for formal mathematical reasoning is a nascent area that integrates insights from formal mathematics, programming languages, and machine learning. We hope this position paper can present coherent perspectives that unite previously fragmented efforts across these fields, fostering discussion, community building, and a clear roadmap for the future.

The narrative in this paper is rooted in the approach of AI and machine learning researchers, emphasizing general-purpose learning algorithms applied to well-defined tasks that can be automatically evaluated using benchmarks. While this paradigm has dominated AI research in recent decades, it has limitations. Mathematicians have explored many ways of using AI in their work, including brainstorming ideas and inspirations, writing assistance, and organizing or searching mathematical literature. Many of these use cases, however, resist straightforward evaluation through benchmarks or automated metrics. Even in areas like theorem proving---where automated evaluation is feasible---performance on benchmarks may not fully capture what human users find meaningful or helpful (Sec.~\ref{subsec:tools}). For example, benchmarks such as LeanDojo~\cite{yang2023leandojo} and PutnamBench~\cite{tsoukalas2024putnambench} often fail to measure how the prover performs on new and evolving formalization projects, such as formalizing the proof of Fermat's Last Theorem~\cite{flt}. This gap underscores the need for human-centered evaluation approaches that draw on insights from human-computer interaction and cognitive science~\cite{frieder2024mathematical,collins2024evaluating}.

\section*{Acknowledgements} 
We gratefully acknowledge Jeremy Avigad, Albert Q. Jiang, Zhaoyu Li, Peter O'Hearn, Daniel Selsam, Armando Solar-Lezama, and Terence Tao for providing valuable feedback on an initial version of this paper.

\clearpage
{\small
\bibliographystyle{unsrtnat}
\bibliography{references}
}

\end{document}